\documentclass[journal]{IEEEtran}
\ifCLASSINFOpdf
\else
\fi

\usepackage{multicol}
\usepackage{times}
\usepackage{epsfig}
\usepackage{graphicx}
\usepackage{amsmath}
\usepackage{amssymb}
\usepackage{algorithmic}
\usepackage[lined,boxed,commentsnumbered]{algorithm2e}
\usepackage{bm}
\usepackage{caption}
\usepackage{cite}
\usepackage{color}
\usepackage[export]{adjustbox}
\usepackage{graphicx}
\usepackage{subcaption}
\usepackage{url}
\usepackage{multirow}
\usepackage[pagebackref=true,breaklinks=true,letterpaper=true,colorlinks,bookmarks=false]{hyperref}

\makeatletter
\newcommand{\mypm}{\mathbin{\mathpalette\@mypm\relax}}
\newcommand{\@mypm}[2]{\ooalign{%
  \raisebox{.1\height}{$#1+$}\cr
  \smash{\raisebox{-.6\height}{$#1-$}}\cr}}
\makeatother

\graphicspath{ {./figs/ADSC_dataset/}{./figs/edge_boxes_iou/}{./figs/fig_1less_step/}{./figs/fig_heatmap/}{./figs/fig_mesh_BB/}{./figs/fig_plane_removal/}{./figs/fig_rgbd_scenes/}{./figs/matched_pix/}{./figs/remaining_figs/} {./figs/updated_3D_map/}{./figs/UW_res/}{./figs/bio_pics/} }

\DeclareMathOperator*{\argmax}{arg\,max}

\newcommand{\rulesep}{\unskip\ \vrule\ }

\begin{document}
%
\title{Locating 3D Object Proposals: A Depth-Based Online Approach}
%
%

\author{Ramanpreet~Singh~Pahwa,
        Jiangbo~Lu,~\IEEEmembership{Senior Member,~IEEE,}
		Nianjuan~Jiang,~\IEEEmembership{Member,~IEEE,}
		Tian~Tsong~Ng,
		Minh~N.~Do,~\IEEEmembership{Fellow,~IEEE}
\thanks{R.~S.~Pahwa, J.~Lu and N.~Jiang are with Advanced Digital Sciences Center (ADSC), Singapore. }
\thanks{Minh~N.~Do and R.~S.~Pahwa are with the Department
of Electrical and Computer Engineering, University of Illinois at Urbana-Champaign, IL, 61801 USA.}
\thanks{T.~T.~Ng is with Institute for Infocomm Research (I$^2$R), Singapore.}
}

%
%

\markboth{IEEE Trans. on Circuits and System for Video Technology}%
{Shell \MakeLowercase{\textit{et al.}}: Bare Demo of IEEEtran.cls for IEEE Journals}
%

\maketitle

\begin{abstract}\label{sec:Abstract}
2D object proposals, quickly detected regions in an image that likely contain an object of interest, are an effective approach for improving the computational efficiency and accuracy of object detection in color images. In this work, we propose a novel online method that generates 3D object proposals in a RGB-D video sequence. Our main observation is that depth images provide important information about the geometry of the scene.  Diverging from the traditional goal of 2D object proposals to provide a high recall, we aim for precise 3D proposals. We leverage on depth information per frame and multi-view scene information to obtain accurate 3D object proposals. Using efficient but robust registration enables us to combine multiple frames of a scene in near real time and generate 3D bounding boxes for potential 3D regions of interest. Using standard metrics, such as Precision-Recall curves and F-measure, we show that the proposed approach is significantly more accurate than the current state-of-the-art techniques. Our online approach can be integrated into SLAM based video processing for quick 3D object localization. Our method takes less than a second in MATLAB on the UW-RGBD scene dataset on a single thread CPU and thus, has potential to be used in low-power chips in Unmanned Aerial Vehicles (UAVs), quadcopters, and drones. 
\end{abstract}

\begin{IEEEkeywords}
Robot vision, depth cameras, object proposals.
\end{IEEEkeywords}

%
\IEEEpeerreviewmaketitle

\begin{section}{Introduction}\label{sec:intro}

\IEEEPARstart{T}{he} rapid development of low-powered Unmanned Aerial Vehicles (UAVs), drones and service robots has introduced a need for automatic detection of interesting objects present in a scene. Such applications may not only assist in navigation of these devices but also help in localizing and identifying the objects that are present in a scene. This paper presents a framework where we integrate depth information captured using a depth sensing device with $2$D color images, build a $3$D global map of the environment and estimate $3$D bounding boxes for the objects of interest that may be present in a given scene by exploiting multi-view scene geometry. 

The idea of $2$D object proposals is in fact not new. It has been presented in the form of unsupervised image segmentation for more than a decade. It is until recently, $2$D object proposal techniques have become popular in object detection systems. Instead of finding precise and exclusive boundaries for objects, modern $2$D object proposal techniques quickly identify regions (potentially highly overlapped) that are very likely to contain an object \cite{cheng2014bing, CPMC, MCG, zitnick2014edge, krahenbuhl2014geodesic, oneata:hal-01021902}. Most importantly, generic object proposals are class independent. This helps in reducing the search space to allow for more sophisticated recognition algorithms and helps in pruning away false positives making the detection step easier \cite{deepbox}. The ability to perceive objects in the scene regardless of their categories is crucial for robots and vehicles, as it can assist the AI agents to understand and explore the real world environment on-the-fly. Recently, Song and Chandraker \cite{song2015joint} present an impressive framework where they combine cues from Structure from Motion (SFM), object detection and plane estimation to estimate rigid $3$D bounding boxes for moving cars in an outdoor real environment. However, their approach requires a prior knowledge of the number of cars present in a given scene. As they use monocular SLAM to estimate camera poses and track $3$D objects in the scene using feature points, their system may suffer in low-texture environments and lack $3$D object dimensions and distances that are critical for actual autonomous navigation.

An effective way to obtain true dimension and distance of $3$D objects is to use depth cameras. The use of depth cameras are ubiquitous nowadays. Due to their popularity, depth camera research has also exploded in the last few years with discoveries of various applications in computer and robot vision \cite{yahav20073d, wilson2010using,  shotton2013real, Biswas, du2011interactive, Song_TCSVt2014}, including object recognition \cite{depth_obj_recog}. Unfortunately, existing object proposal solutions designed for RGB images are sub-optimal for depth camera video input, as no depth information is used and no dense temporal consistency is imposed. Scene geometry should be used by a robot to its advantage 
and the robot can maneuver around the scene to gain critical information about the objects that may be present in the scene \cite{Xu_2015_Reconstruction}. 

In this work, we propose an online $3$D object proposal technique for RGB-D video input of a static scene with the main focus on precision. We use the per-frame result given by existing $2$D proposal techniques as our input. Most of the $2$D object proposal methods focus on high recall and output several object proposals around an object of interest. Selecting the best object bounding box among these object proposal candidates is non-trivial. We leverage on segmentation cues provided by depth information and aggregate them over consecutive frames in $3$D by estimating the camera poses using RGB-D SLAM. By doing this, we achieve high precision with light computational overhead. Some recent works use deep learning in a supervised manner to incorporate segmentation cues right from the start for $2$D object proposals and object detection \cite{deepbox, NIPS2015_Faster_RCNN}. These methods can achieve online object recognition albeit with the use of heavy parallel processing and expensive GPUs. While appreciating the technical elegance of such approaches, we adopt a more practical approach with the objective of providing an efficient and versatile system that can work with any off-the-shelf $2$D proposal techniques. In fact, one major distinction of our method from these existing single color image based methods is that we use a RGB-D video sequence. Depth information helps to localize and segment objects more easily with access to scene geometry, which still remains a challenge for color-only approaches. The proposed method can also segment occluded objects in a single frame by taking advantage of the frames in the video sequence. This allows consolidating information across frames jointly to segment out the occluded objects correctly, without placing a burden on the per-frame object segmentation process. Overall, our approach is both effective and efficient to find objects of interest in $3$D by using a RGB-D video sequence.

More specifically, this work focuses on indoor, static scenes containing a major supporting plane. However, it can be easily extended to remove multiple supporting planes. The objects are allowed to occlude each other and partially seen in some images as long as their complete view is covered over a consecutive number of frames. Note that we do not assume their distance to the camera.

We make the following contributions: 
\begin{itemize}
\item  We integrate depth information with state-of-the-art $2$D object proposal techniques to improve object proposals per frame. 

\item During the scene capturing process, we exploit the indoor scene geometry to automatically remove any supporting planes such as tables which are not a part of the objects of interest.

\item  We use the camera pose estimated by a depth based SLAM technique to efficiently register the frames to a global point cloud and output multi-view consistent $3$D object proposals as the camera moves around in the scene. 
\end{itemize}

To showcase our results, we perform density-based $3$D clustering on the top-ranked points, and display our proposed $3$D objects bounded by tight $3$D bounding boxes. One key aspect of our approach is that it is efficient using only a single thread CPU, so our system has a good potential of being integrated on top of existing depth based SLAM methods. 

Our paper is structured as follows. In Section \ref{sec:related_work}, we review other works that are related to our research topic. In Section \ref{sec:overview}, we provide a brief overview of our problem formulation. Section \ref{sec:2D_heatmap} and \ref{sec:3D_fusion} present the proposed $2$D depth based filtering and $3$D fusion and refinement steps in detail, highlighting our contributions and observations at each stage. We report our results and comparison with other state-of-the-art methods in Section \ref{sec:results}. Finally, in Section \ref{sec:conclusion} we conclude this paper and discuss future research directions.
\end{section}
\begin{figure}[t]
  \includegraphics[trim={0.0cm 17cm 4cm 0cm}, clip, width=0.45\textwidth]{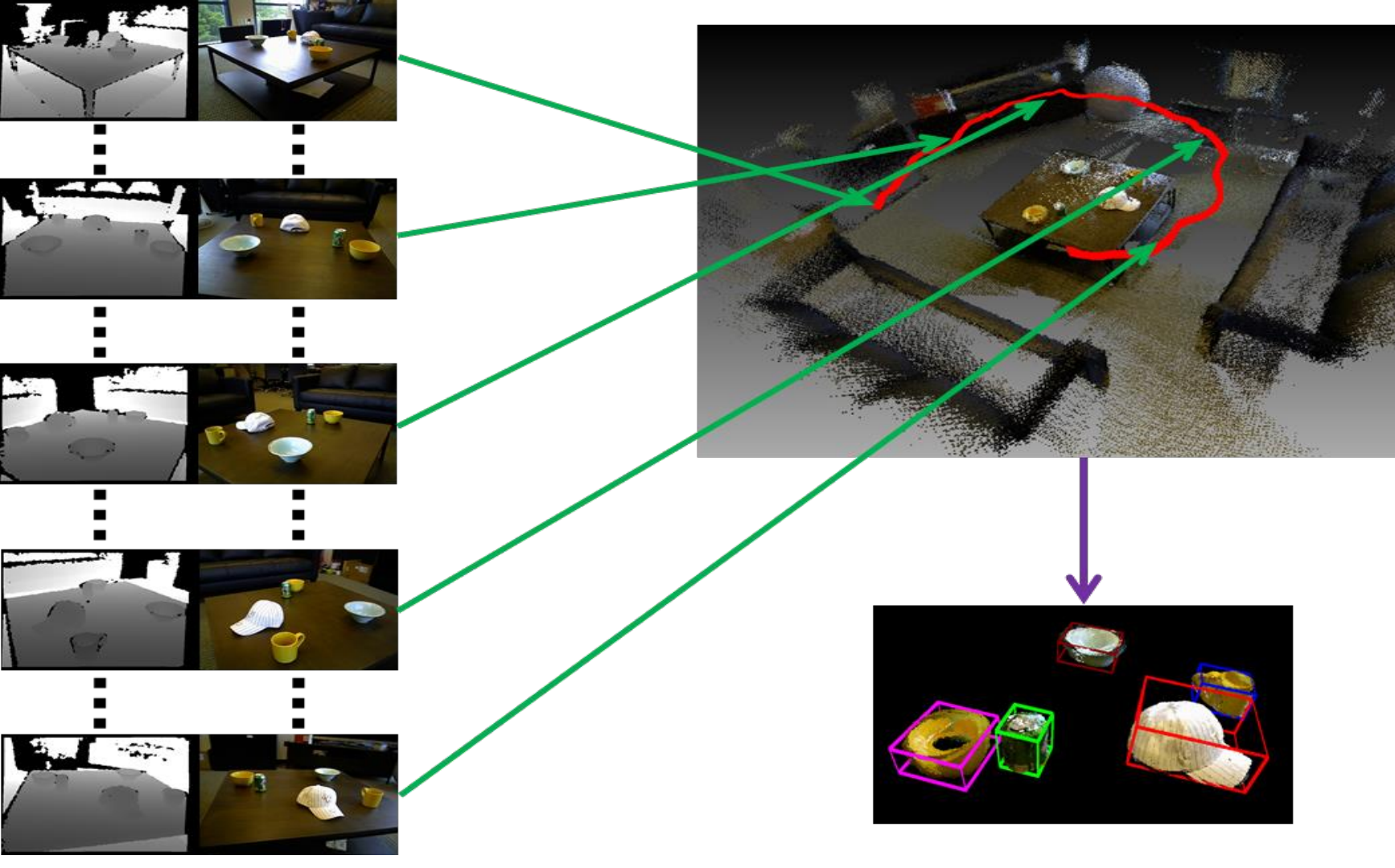}
\caption{Scene data is captured using a RGB-D camera. The aligned color and depth images are used for camera pose estimation (illustrated by the red trajectory). Our approach uses this information along with any generic $2$D object proposals to fuse and filter the data in $3$D and output precise $3$D object proposals (denoted by colored $3$D bounding-boxes).}
\label{fig:our_traj_demo}
\end{figure}
\begin{section}{Related Work}\label{sec:related_work}
\textbf{Object proposals:} 
A lot of work has been done recently on $2$D object proposals. Traditionally, sliding windows are used along with a detector to identify objects in a given scene. Many of the state-of-the-art techniques in object detection have started using a generic, object-class agnostic proposal method that finds anywhere between $100$-$10,000$ bounding boxes in an image. These areas are considered to have the maximum likelihood to contain an object in them. Such methods vary widely from using linear classifiers, BING \cite{cheng2014bing}, graph cuts, CPMC \cite{CPMC}, graph cuts with an affinity function \cite{hoiem_PAMI2014}, normalized cuts, MCG \cite{MCG} to using  random forests, edge-boxes \cite{zitnick2014edge}, geodesic transform \cite{krahenbuhl2014geodesic} and superpixels \cite{oneata:hal-01021902}. Ren \emph{et~al.} \cite{NIPS2015_Faster_RCNN} use deep learning in a supervised manner to find $2$D object proposals and perform object detection simultaneously. 

Hosang \emph{et~al.} \cite{hosang2015makes} provide a systematic overview of these object proposal techniques and tested these state-of-the-art algorithms. They demonstrate that most of these algorithms have limited repeatability. Even changing one pixel exhibited markedly different outcomes. They found edge-boxes and geodesic proposals to give the best results. Based on these observations, we decided to use edge-boxes \cite{zitnick2014edge} as it is reported to be significantly faster and slightly more accurate than geodesic object proposals.
\begin{figure*}[t!]
\centering
\includegraphics[trim={0.5cm 37cm 18cm 0.5cm},clip,width=1.00\textwidth]{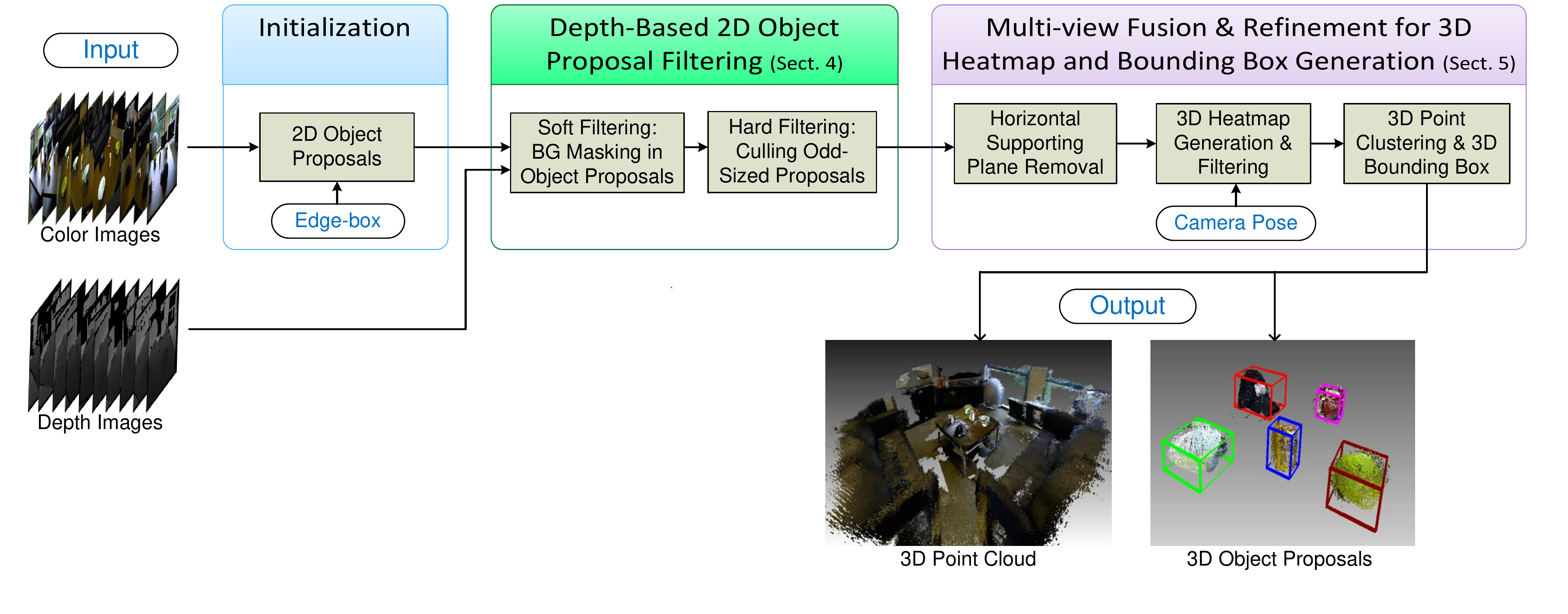}
\caption{Our proposed framework for $3$D object proposals. We utilize a generic $2$D object proposal scheme such as Edge-boxes \cite{zitnick2014edge} to obtain $2$D object proposals. These proposals undergo robust filtering and culling based on scene geometry obtained from corresponding depth images. A horizontal supporting plane is identified and removed in each image and a $3$D heatmap is sequentially obtained. Neighboring frames are fused together using camera poses to obtain $3$D point cloud and a consolidated heatmap. Top $3$D points are selected from the consolidated $3$D heatmap based on a ranking scheme and fast $3$D clustering is performed on these points to obtain $3$D object proposals.} 
\label{fig:flowchart}
\end{figure*}

\textbf{Simultaneous localization and mapping (SLAM) with scene geometry:} 
Choudhary \emph{et~al.} \cite{choudhary2014SLAM} propose an online object discovery approach that extends standard SLAM to utilize the discovered objects for pose estimation and loop closures. They use $3$D feature descriptors, CSHOT \cite{CSHOT_2011}, to find the camera pose for every frame. They use $3$D voxels to roughly segment out objects and use them as landmarks for loop closure.

Song and Chandraker \cite{song2015joint} argue that the structure from motion (SFM) cues ($3$D points and ground plane) and object cues (bounding boxes and detection scores) complement each other. The former is accurate for nearby environments while the latter is more accurate for objects that are far away. They combine these two approaches and propose a framework for $3$D object localization. 

Pillai and Leonard \cite{pillai2015monocular} developed a monocular SLAM-aware system that used temporal information to output multi-view consistent object proposals using efficient feature encoding and classification techniques. They only utilize color information of the scene. Their approach works well if the objects are far apart so that the monocular SLAM method can provide a reliable semi-dense depth estimate for each view. However, their approach may fail on more crowded environments (lots of objects) or when objects are cluttered together in a small region. Using depth information enables us to handle these situations as we can exploit the $3$D scene geometry to identify regions of interests, redundant vertical and horizontal planes and cluster them separately. We perform our evaluation on the same dataset for a direct comparison. We demonstrate that our results show significant improvement when compared with theirs. Another, and perhaps the most significant difference in our approach is that we use depth images along with color images to output $3$D object proposals while they compute $2$D object proposals only.

\textbf{Single-view RGB-D object proposals:} 
\cite{fouhey2012object,janoch2013category,gupta2014learning} found depth based proposals to be inferior to color based proposals. They conclude that depth and color combined together can potentially give worse results when compared to using color alone due to the inherent noise present in depth images. 

Recently, a number of hybrid $2.5$D approaches have been proposed, which transform depth input into more useful information such as ``height above ground'', ``angle with gravity'' and ``surface normals'' \cite{gupta2014learning} to be used together with color information per pixel for object proposals. Fouhey \emph{et~al.} \cite{fouhey2012object} proposed an adaptive approach where they fuse depth with color using a maximum likelihood estimation technique. They demonstrate that using a weighted scheme gave them better results than using just color or using color and depth without adaptive weights. We utilize the depth information to estimate the plane normal and horizontally planar points. This enables us to transform our $3$D bounding boxes to align vertically with the global point cloud. \textbf{Chen \emph{et~al.} \cite{chen_arxiv, chen_nips15} introduced a MRF formulation with hand-crafted features for a cars, pedestrian and cyclists detection in street scenes. They use pre-learned box templates of different sizes from trained data and use estimated point cloud per frame to generate $3$D object proposals. On the contrary, our approach is designed towards  indoor scenes with no prior training or knowledge about objects we detect in the scenes. Song and Xiao \cite{song_arxiv, song_eccv14} train exemplar-SVMs on synthetic data on hundreds of rendered views. They slide these exemplars in $3$D space to search for objects by densely evaluating $3$D windows. They perform object detection on pre-trained objects per frame and require gravity direction to be known. However, we leverage on scene information by aggregating information across video frames and output $3$D object proposals without having any prior information about the objects present in a given scene.}
\end{section}

\begin{figure}[t]
\centering
\includegraphics[trim={0.0cm 15cm 7.5cm 0cm},clip,width=0.45\textwidth]{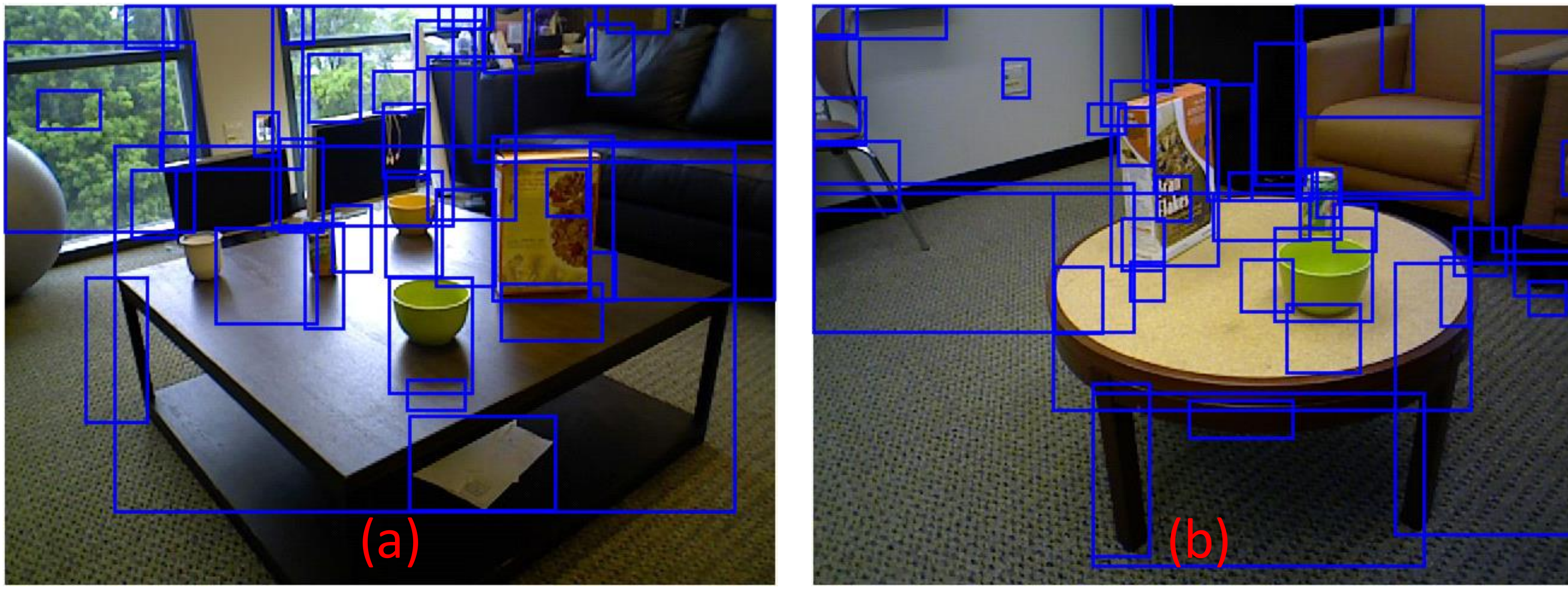}
\caption{Top ranked object proposals are displayed after Non-Maximum Suppression (NMS). Object proposals are displayed in decreasing order of their confidence. Object proposal with the highest confidence is displayed first. Any object proposal that shares more than $10\%$ region is not shown. Using a simple NMS approach is unable to select the best object proposals.} 
\label{fig::bb_2d_jaccard}
\end{figure}

\begin{section}{Overview of the Proposed Approach}\label{sec:overview}

We start by giving an overview of the proposed algorithm. At high level, the proposed algorithm is designed to fuse the depth information with the generic object proposals obtained from using color images in an indirect manner. This enables us to exploit using $3$D geometry of the scene without impacting the results due to noisy or unavailable depth information at various image pixels. Figure \ref{fig:our_traj_demo} illustrates the basic setup of the problem studied in this paper and an example result obtained by our algorithm. Our framework consists of a few novel modules in its pipeline (as shown in Fig.~\ref{fig:flowchart}), and they will be presented in detail next. First, we introduce the initialization process in this section. Sec.~\ref{sec:heatmap_creation} presents the representation for the local $2$D heatmap. Sec.~\ref{sec:soft_filtering} and \ref{sec:hard_filtering} describe our depth based filtering process. Sec.~\ref{sec:3D_fusion} discusses how we obtain a global $3$D heatmap and our final $3$D object proposals.

Before presenting the main components of our approach, we first introduce the initialization process with $2$D object proposals. We collect $N$ video frames per scene using a RGB-D camera. Every $i^{th}$ video frame consists of a color image $\bm{I}_{i}$, a depth image $\bm{Z}_{i}$, and the pose of the camera $\bm{P}_{i}$, using a depth based SLAM method such as Dense Visual SLAM \cite{kerl2013dense}. The camera pose contains the rotation and translation measurements of the camera in the world coordinate frame of reference: $\bm{P}_{i} = [\bm{R}_i,\bm{t}_i]$. The RGB-D camera is assumed to be pre-calibrated. The camera intrinsic parameters - the focal length in $x$ and $y$ directions and optical center are denoted by $f_x, f_y, [c_x, c_y]$, respectively. Together, these are represented by the intrinsic calibration matrix $\bm{K}$. 

We use a generic $2$D object proposal technique (edge-boxes \cite{zitnick2014edge}) to obtain $M$ $2$D object proposals per frame. Edge-boxes provide a $1\times 5$ measurement vector per bounding box:
\begin{equation} \label{Eq:EB}
\bm{e}^{j}_{i} ={[x^j,y^j,w^j,h^j,c^j]_i}, \enskip j \in {1, \hdots, M}, \enskip i \in {1, \hdots, N}. 
\end{equation}
where, $[x,y]$ denote the top-left pixel coordinate of the bounding box, and $w,h$ refer to the width and the height in $x$ and $y$ directions respectively. $c$ corresponds to the confidence value of the bounding box which is proportional to the probability of that bounding box containing an object in it. A few top ranked $2$D proposals are shown in Fig.~\ref{fig::bb_2d_jaccard}. These $2$D object proposals per image are treated as an input for our framework.
\end{section}

\begin{section}{Depth-Based 2D Object Proposal Filtering}\label{sec:2D_heatmap}

In this section, we describe how we use depth information per frame to improve existing $2$D object proposal techniques. As we only look at a single frame here, the subscript $i$  is dropped for the clarity sake from the single view formulation.

\begin{subsection}{2D Weighted Heatmap for Pixelwise Objectness}\label{sec:heatmap_creation}

One problem with using object proposals is the redundancy of majority of the proposals. The object proposal techniques are usually designed for high recall and Jaccard index (IoU). This results in several object proposals in image areas where an object might be located. Object recognition techniques are used next on each of these bounding boxes to detect and identify if they contain an object. We ask a question: is it possible to quickly improve or reject some of these proposals using scene geometry without performing expensive object detection and recognition techniques?

To select the best possible proposal, techniques similar to non-maximum suppression (NMS) are generally used. Object proposals with the highest confidence locally are chosen, while any neighboring proposals that share a common region with the locally chosen proposal are rejected. However, there is no guarantee that the object proposals with higher confidence will actually be the best fitting object proposals for a given scene as shown in Fig.~\ref{fig::bb_2d_jaccard}. \cite{pillai2015monocular} also observed that object proposals techniques often fail to find good $2$D boundaries of objects present in a video scene due to motion blur. For example, none of the top $1,000$ proposals in Fig.~\ref{fig::bb_2d_jaccard}(a) resulted in a good bounding box for the yellow bowl (IoU $> 0.5$). The technique described in this paper is able to overcome these motion blur based problems and find good $3$D and $2$D object proposals in such cases well.

Thus, instead of deciding right away the best 2D proposals in a given scene, a weighted approach is used to create a \textit{heatmap}. A heatmap is a two-dimensional real-valued matrix indicating the probability of each image pixel $[u,v]$ to be occupied by an object. We consolidate the confidence of each pixel by summing over all the object proposals $\{{\bm{e}^{j} = [x^j,y^j,w^j,h^j,c^j]}\}^M_{j = 1}$ for a given image. We denote this heatmap as ``baseline heatmap'' $\hat{\bm{H}}$, and it is obtained by:
\begin{equation}
\hat{\bm{H}}[u,v] = \sum_{j = 1}^M c^j\delta_o^{j} [u,v]
\end{equation}
\begin{equation}
\delta_o^{j}[u,v] = \begin{cases} 
1 & \text{if  } u \in [x^j, x^j + w^j], \text{  } v \in [y^j, y^j + h^j] \\ 
0 & \text{otherwise} \;,  \\ \end{cases} 
\end{equation}
where $c^j$ denotes the confidence of the $j^{th}$ object proposal, and $\delta_o^{j}[u,v]$ is a binary filter that checks if the pixel $[u,v]$ is contained in the current object proposal ${\bm{e}^{j}}$. An example of a baseline heatmap is provided in Fig.~\ref{fig:heatmap_comparison}(c).
\begin{figure}[t!]
\centering
\includegraphics[trim={0.0cm 7cm 0cm 0cm},clip,width=0.45\textwidth]{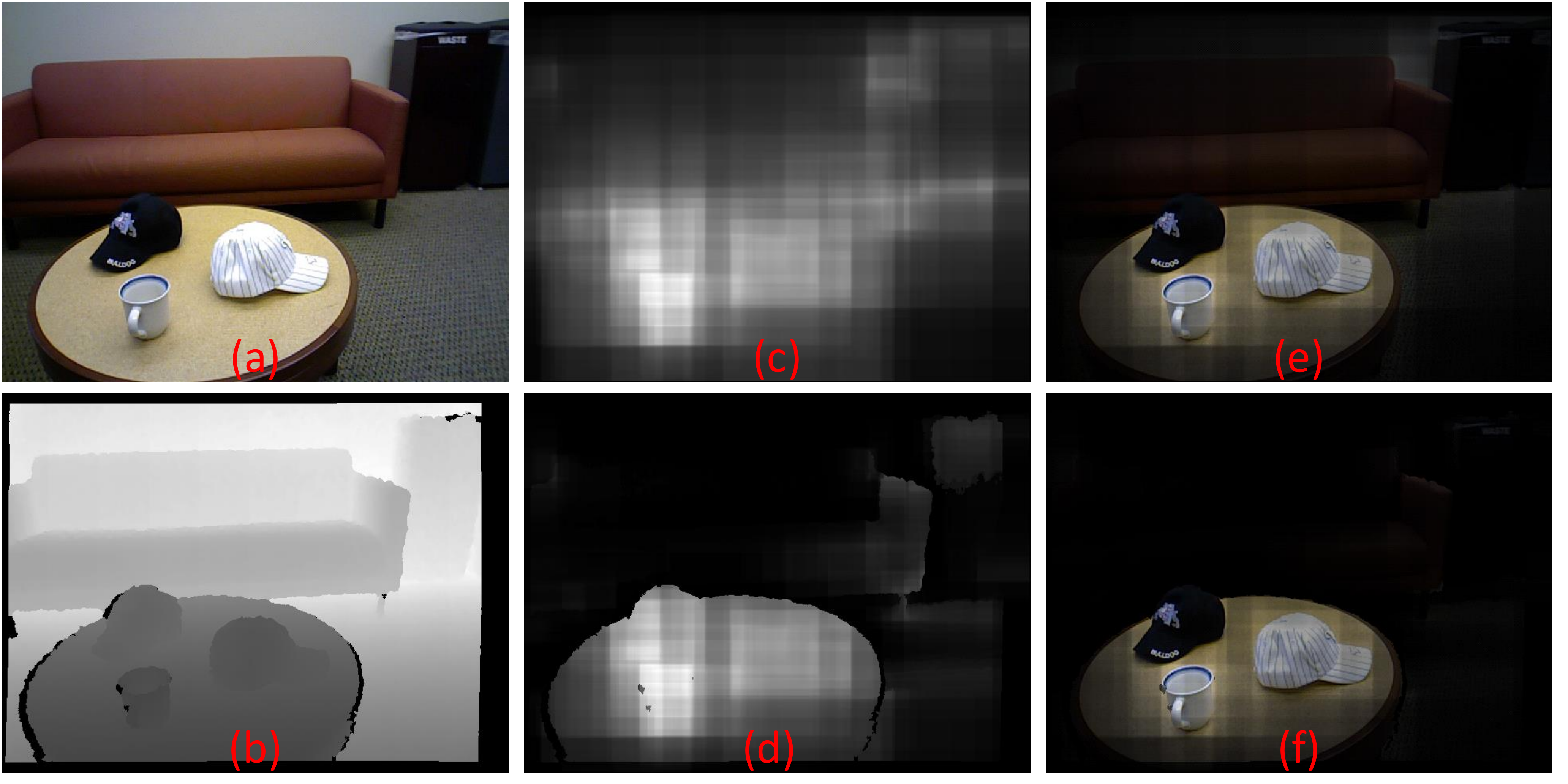}
\caption{(a) Color image (b) Depth image (c) Baseline heatmap $\hat{\bm{H}}$ (d) Improved heatmap $\bm{H}_{2D}$ (e) Baseline heatmap overlaid on color image (f) Our heatmap overlaid on color image. We show a comparison between the baseline heatmap $\hat{\bm{H}}$ and our improved heatmap $\bm{H}_{2D}$. Our depth-assisted heatmap is able to filter out the background from the objects of interest using depth based statistics for every object proposal.} 
\label{fig:heatmap_comparison}
\end{figure}
There are various advantages of using a heatmap based approach. Firstly, it is computationally fast and extremely simple to implement. The baseline heatmap can be computed in constant time using integral images \cite{Crow_integral_img}, regardless of the size of bounding boxes. Secondly, the heatmap provides a much more fine-grained, per-pixel objectness map rather than an array of bounding boxes per frame. This makes it easier to transfer this information to a $3$D global heatmap. Thirdly, the heatmap is an accumulation of statistics computed from all bounding boxes. It turns out that sometimes no bounding box may cover an object properly. In such cases, the heatmap obtained from all these bounding boxes may result in a far more intelligent boundary of the object's location in the scene. Thus, instead of matching bounding boxes across video frames one by one, we project the entire heatmap and consolidate the confidence values of the $3$D points in the global world coordinate frame of reference.  

The baseline heatmap provides a good idea where the objects of interest might be located in the scene. However, the baseline heatmap is still rather coarse as seen in Fig.~\ref{fig:heatmap_comparison}(c). This is because object proposals tend to enclose a lot of non-object regions such as texture on the wall and floor or sides of the table as seen in Fig.~\ref{fig:heatmap_comparison}(e). To resolve this issue, a depth based two-step filtering process is performed to refine these initial $2$D proposals: \textit{soft filtering} and \textit{hard filtering}.
\end{subsection}

\begin{subsection}{Soft Filtering: Background Masking in Object Proposals}\label{sec:soft_filtering}

Soft filtering refers to making a binary decision for each pixel in an object proposal. Every pixel of an object proposal region  is classified into a foreground or background pixel by performing a quick segmentation using depth information. Let $Z_{min}$ and $Z_{max}$ be the minimum and maximum depth range of the current object region. We compute its mean: $Z_\mu = \frac{Z_{min}+Z_{max}}{2}$. As our system is geared towards online applications, we avoid employing more complex segmentation techniques such as mean-shift or k-means. An object proposal is considered as containing background parts if the difference $\Delta Z (= Z_{max} - Z_{min})$ is above a certain threshold $\epsilon_{\Delta}$. 

For such object proposals containing background portions, a pixel $[u,v]$ is classified as a background pixel if the pixel's depth value $\bm{Z}[u,v]$ is more than $Z_\mu$ and vice versa as given in Eq.~\ref{eq::bg_fg_classification}. The background pixels in each bounding box proposal are assigned zero confidence value while the confidence value for foreground pixels is retained: 
\begin{equation}\label{eq::bg_fg_classification}
\delta_s^{j} [u,v] = \begin{cases} 
0 & \text{if  } \bm{Z}[u,v] > Z^j_{\mu} \quad \text{and} \quad \Delta Z^j >  \epsilon_{\Delta} \\ 
1 & \text{otherwise} \;, \\ \end{cases}
\end{equation}
where $\epsilon_{\Delta}$ is a small threshold. The thresholds are very generous and only mask off background pixels when there is a high probability for a proposal to contain the background. Figure~\ref{fig:3Dedgebox_proposal} demonstrates the effect of this soft filtering method. The blue and magenta bounding boxes represent the regions that are detected to contain background parts and undergo background masking. The red and green bounding boxes represent regions with $\Delta Z <  \epsilon_{\Delta}$, and do not undergo this soft filtering process as they do not contain any significant background. 
\end{subsection}  
\begin{figure}[t!]
\centering
\begin{subfigure}[b]{0.45\textwidth}
\includegraphics[trim={0.0cm 0cm 15cm 0cm},clip,width=\textwidth]{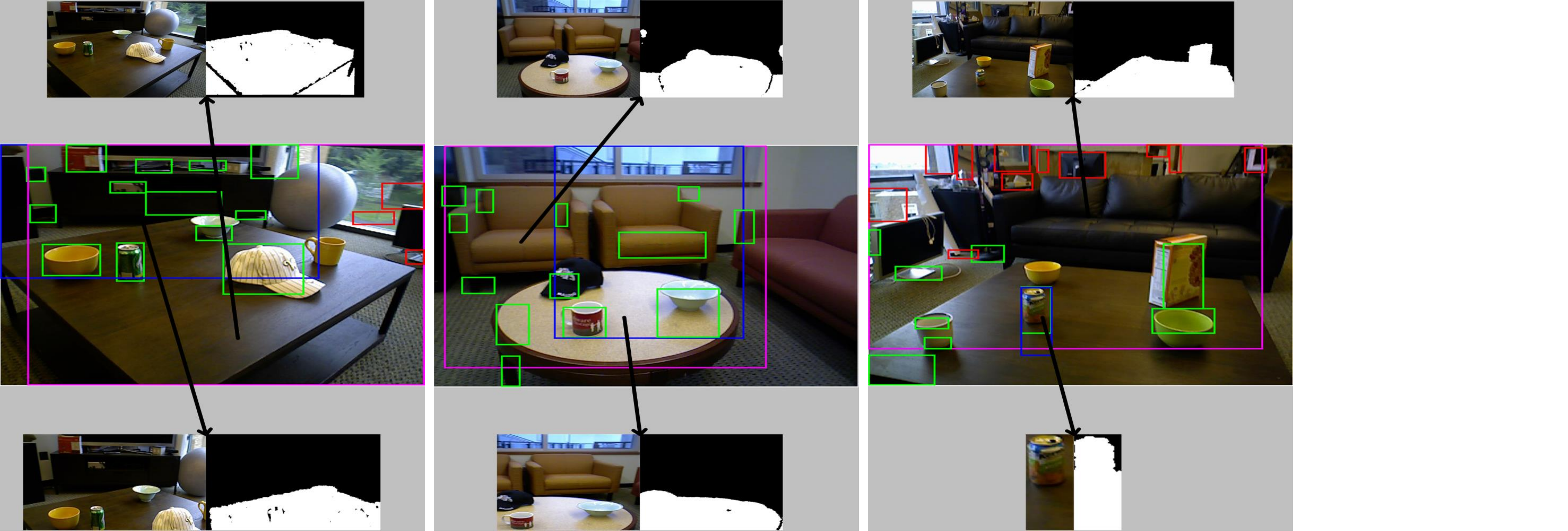}
\end{subfigure} 
\caption{Depth based filtering of $2$D object proposals. We only display a handful of object proposals for clarity purposes. In the soft filtering process, the blue and magenta bounding boxes are detected to contain background and undergo background suppression while the red and green bounding boxes are detected to contain no major background. After the background masking, we check the actual size of each bounding box using depth information. The red and magenta bounding boxes are identified to be either too small or too big in size or do not contain depth information. These $2$D object proposals are discarded. The blue regions, after background suppression, and green regions are detected to potentially contain good information and are accepted as valid object proposals.} 
\label{fig:3Dedgebox_proposal}
\end{figure}  

\begin{subsection}{Hard Filtering: Culling Odd-Sized Object Proposals}\label{sec:hard_filtering}

Hard filtering refers to making a binary decision, accept or reject, for every object proposal of an image. The $2$D bounding boxes given by  object proposal techniques can vary from anywhere between $2\times2$ pixels to spanning the entire image. To reduce the number of irrelevant bounding boxes, some approaches \cite{pillai2015monocular} ignore any small bounding boxes less than $20\times20$ pixels or any relatively large bounding boxes. However, sometimes this will result in losing important information when the object is located at a distance from the camera or when the camera is zoomed onto an object momentarily. 

To overcome this problem, the size statistics of each bounding box are computed using the depth information available. An object proposal is discarded if it is estimated to be outside the desired range as follows,  
\begin{equation}\label{eq::patch_size_eq}
\delta_h^{j} = \begin{cases} 
        \multirow{2}{*}{0} & \text{if  }  \frac{w^j*Z_{\mu}^j}{f_x},\frac{h^j*Z_{\mu}^j}{f_y} <  \epsilon_{min} \text{    or   }\\
        & \quad \frac{w^j*Z_{\mu}^j}{f_x},\frac{h^j*Z_{\mu}^j}{f_y} >  \epsilon_{max} \\
        1 & \text{otherwise} \;, \\ \end{cases}
\end{equation}
where $[\epsilon_{min},\epsilon_{max}] = [2cm,1m]$, and $w^j, h^j,Z_{\mu}^j$ correspond to width, height and mean depth of the foreground object in the $j^{th}$  proposal. $f_x$ and $f_y$ are the known focal length parameters in the $x$ and $y$ direction. Specifically, any proposal whose approximate cross-section size is bigger than $1m \times 1m$ or smaller than $2cm \times 2cm$ is rejected. Note that a proposal of size $3m \times 1cm$, such as a long stick, is still considered to be a desirable object and accepted as a valid proposal. The parameters $\epsilon_{min}$ and $\epsilon_{max}$ are empirically set in this paper, but they can be adapted to best fit different RGB-D sensing cases. Object proposals that do not contain any depth information are also rejected. This usually happens when the regions inside these bounding boxes are too far from the depth camera. In Fig.~\ref{fig:3Dedgebox_proposal}, the red and magenta bounding boxes represent proposals that are rejected based on our hard filtering process. Meanwhile, the green and blue bounding boxes represent proposals that are accepted as valid proposals.

Thereafter, the confidence for each object proposal window is updated pixel by pixel, and our improved weighted $2$D heatmap is computed for every pixel $[u,v]$ as follows: 
\begin{equation}\label{eq::2d_heatmap}
\bm{H}_{2D}[u,v] = \sum_{j = 1}^M \delta_h^{j} \cdot \delta_s^{j} [u,v] \cdot c^j\delta_o^{j}[u,v]  \;.
\end{equation}

In summary, our depth-based filtering approach reduces the computation required and improves the precision of our object proposals as the irrelevant object proposals are filtered out. A baseline heatmap (without the depth based filtering process) and our weighted heatmap are shown in Fig.~\ref{fig:heatmap_comparison}(c,d). The objects of interest (cup and caps) stand out in our heatmap. Additionally, the background is correctly detected and assigned low confidence as shown in Fig.~\ref{fig:heatmap_comparison}(f).
\end{subsection}
\end{section}

\begin{section}{Multi-view Fusion and Refinement for a Global 3D Heatmap}\label{sec:3D_fusion}

In this section we describe our second main contribution -- how a sequence of weighted $2$D heatmaps $\{\bm{H}_{2D}\}$ is fused in $3$D using depth information and camera pose efficiently. 

Let $\bm{x}_w = [x,y,z]^\intercal$ be a $3$D point in the world coordinate frame. It can be projected onto a $2$D camera plane, and we denote the projected $2$D point by $\bm{x}_c = [u,v]^\intercal$ and compute it as $\begin{bmatrix} \bm{x}_c\\ 1 \end{bmatrix}  = \lambda \bm{K} \bm{P} \begin{bmatrix} \bm{x}_w\\ 1 \end{bmatrix}$. Here $\lambda$ is a proportionality constant, while $\bm{K}$ and $\bm{P}$, as defined earlier, denote the intrinsic calibration matrix and the camera pose, respectively. For notational convenience, we represent this projection onto the camera plane as a function $\pi$: $\bm{x}_c  = \pi( \bm{K}, \bm{P}, \bm{x}_w)$.

Similarly, a $2$D pixel $\bm{x}_c$ can be projected and transformed onto the world coordinate frame by using its depth value $\bm{Z}[\bm{x}_c] $:
\begin{eqnarray}
\bm{x}_p  = \begin{bmatrix} \frac{\bm{Z}[\bm{x}_c]}{f_x}(u-c_x) \\ \frac{\bm{Z}[\bm{x}_c]}{f_y}(v-c_y) \\ \bm{Z}[\bm{x}_c]  \end{bmatrix} , \qquad  \bm{x}_w  = \bm{R}^\intercal\bm{x}_p  -\bm{R}^\intercal\bm{t} \;,
\end{eqnarray}
where $\bm{x}_p$ is the $3$D point in the camera frame of reference. It is then transformed to the world coordinate frame using the camera pose $\bm{P}$. We define the projection and transformation as a function $\pi^{-1}$:
\begin{equation}\label{eq::project_2D_to_3D}
\bm{x}_w  = \pi^{-1}( \bm{K}, \bm{P}, \bm{x}_c, \bm{Z}[\bm{x}_c] ) \;.
\end{equation}
%

\begin{subsection}{Horizontal Supporting Plane Removal}\label{sec:plane_removal} 

After the weighted $2$D heatmap $\bm{H}_{2D}$ is computed for the current frame, the image pixels are projected onto $3$D using Eq.~\ref{eq::project_2D_to_3D}. The $3$D points are registered with the global point cloud $\bm{H}_{3D} \in \mathbb{R}^ {N\times 8} : [x,y,z,r,g,b,c,f]$. The first three components $[x,y,z]$ represent the global $3$D location of the point $\bm{x}_w$. The next three components $[r,g,b]$ represent its color. $c$ records the consolidated heat value (confidence) of the $3$D point. Finally, $f$ denotes the number of times (frequency) that a point has been seen in RGB-D frames. We initialize the $3$D heatmap $\bm{H}_{3D}$ for the first frame with unit frequency for all the valid image pixels. 

The next task is to estimate dominant horizontal supporting planes, with the motivation that they are usually not objects of interest in a given scene. Given three $3$D points $\{\bm{x}_1,\bm{x}_2,\bm{x}_3\}$, a $3$D plane ${\bm{p}} = [n_x,n_y,n_z,b]^{\intercal}$  passing through these three points is computed as follows,
\begin{align*}
\bm{n} & = \begin{bmatrix}n_x \\ n_y \\ n_z \end{bmatrix} =  \frac{(\bm{x}_2 -\bm{x}_1) \times (\bm{x}_3 -\bm{x}_1)}{||(\bm{x}_2 -\bm{x}_1) \times (\bm{x}_3 -\bm{x}_1)||_2} \nonumber \\
b & =  -\bm{n}^{\intercal}\bm{x}_1 \;.
\end{align*}

A $3$D point, ${\bm{x}_i}$, is considered lying on the plane ${\bm{p}}$ if
\begin{equation} \label{eq::plane_eq}
\left|{\bm{p}}^{\intercal}\cdot \begin{bmatrix}{\bm{x}_i}\\ 1 \end{bmatrix}\right|< \epsilon_{{\bm{p}}} \;,
\end{equation}
where $\epsilon_{{\bm{p}}}$ is a small threshold to account for noisy data. 
The $3$D points that lie on this plane are considered inliers and the points that exceed the threshold are considered as outliers. The inliers for the $j^{th}$ detected plane ${\bm{p}^j}$ are denoted as ${\bm{S}^j}$. 

One can classify each pixel into different segments using region classifiers and further refine them using shape priors \cite{guo_hoiem}. However, such techniques are not designed to be real time. In contrast, our method quickly identifies the most dominant plane present in the heatmap using RANSAC. Three neighboring points are selected in a given frame to estimate a plane passing through these points. Then, inliers and outliers are computed for the plane using Eq.~\ref{eq::plane_eq}. This process is repeated $10,000$ times to find the top five distinct dominant planes. The top planes for a sample frame in the order of decreasing number of inliers are shown in Fig.~\ref{fig::plane_removal}(c-f).

\begin{figure}[t!]
\centering
\includegraphics[trim={0.0cm 11cm 11cm 0cm},clip,width=0.45\textwidth]{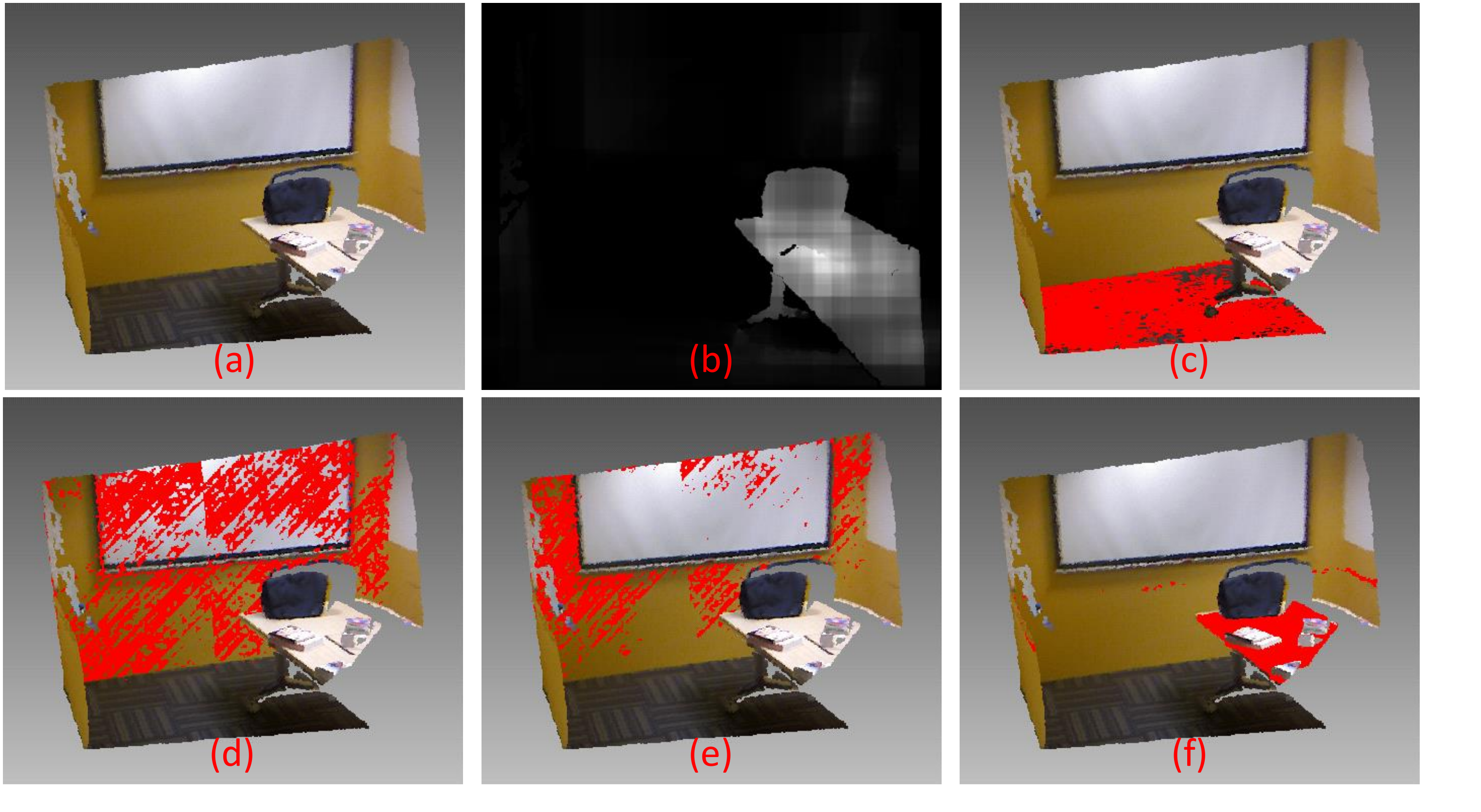}
\caption{Our plane removal process takes into account the heatmap of the scene. A sample $3$D point cloud  and its weighted heatmap is shown in (a) and (b) respectively. We display the most dominant planes in order of number of inliers in (c)-(f). The plane in (f) is selected for plane removal as its heatmap based confidence is highest among top ranked planes.}  
\label{fig::plane_removal}
\end{figure}

To improve the object localization, we aim to separate the objects of interest from a plane that may be present in the top ranked points. As shown in Fig.~\ref{fig::plane_removal}(b), objects of interest often lie on a supporting plane such as the table underneath the objects. This supporting plane is often contained in the $2$D object proposals even after our depth based filtering process. Since a scene capture can start with a camera angle in any direction, we do not make any underlying assumptions about the direction and location of the horizontal planes in a given scene. We observed that using the RANSAC based technique to find the plane with most inliers often resulted in selecting the horizontal plane containing the floor or vertical plane containing room walls. Due to complicated indoor lighting conditions, the table color may also vary significantly from pixel to pixel. Hence, using a color based technique may not always be successful in finding our plane of interest. In addition, it is challenging to know how many dominant redundant planes there might be in a given scene. Thus, removing all the dominant planes is also not an ideal solution.
\begin{figure*}[t!]
    \begin{subfigure}[b]{0.195\textwidth}
		\includegraphics[trim={0.0cm 16cm 14cm 0cm},clip,width=\textwidth]{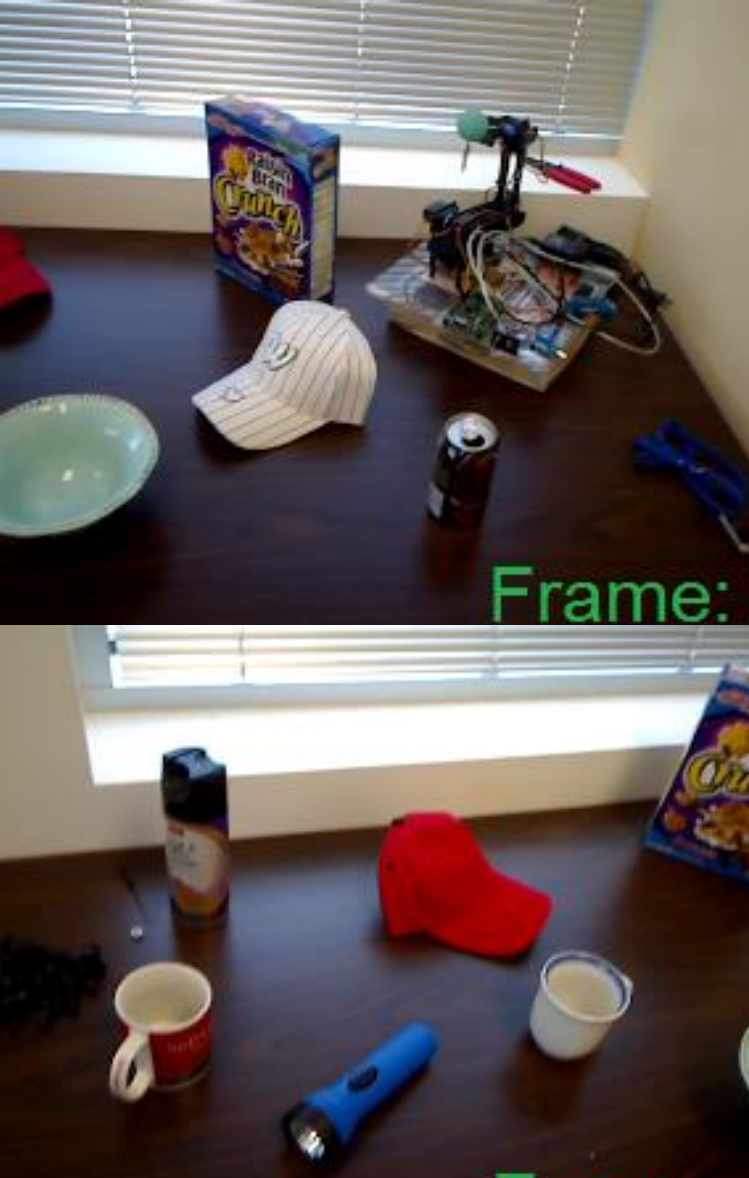}
		\caption{Color image \\ \hspace{\textwidth} }
	\end{subfigure}  
	\begin{subfigure}[b]{0.195\textwidth}
		\includegraphics[trim={0.0cm 16cm 14cm 0cm},clip,width=\textwidth]{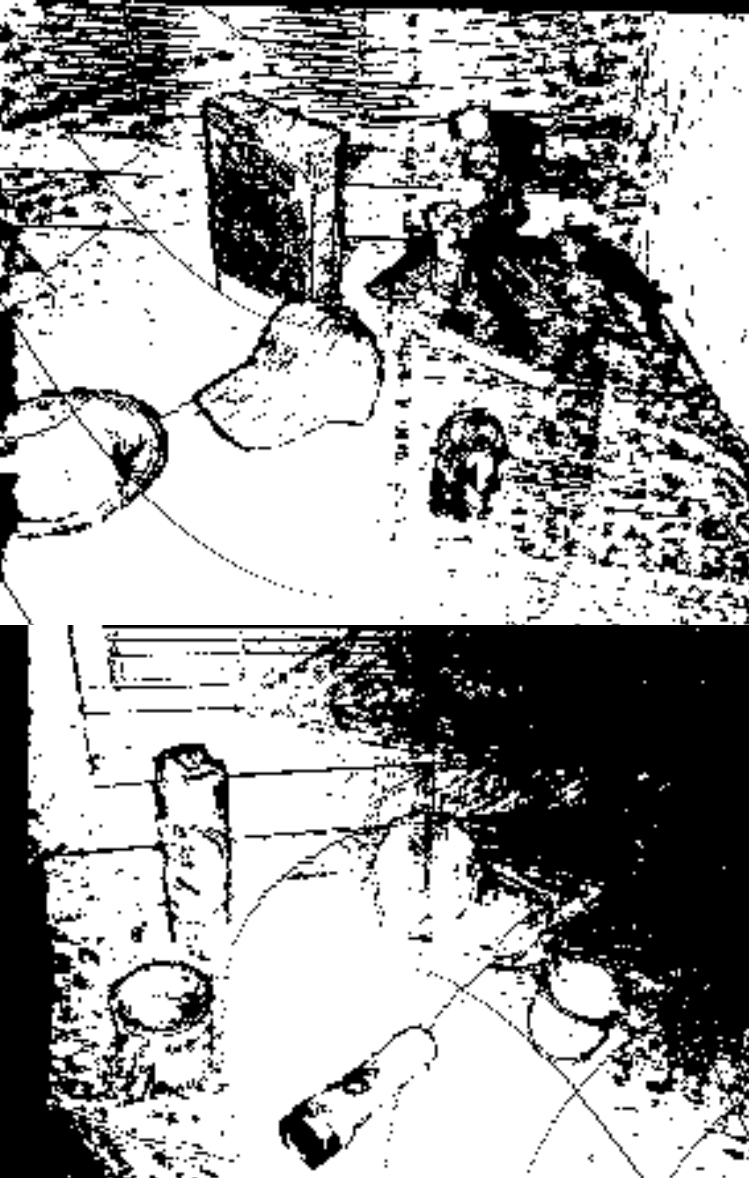}
		\caption{Matched pixels \\ \hspace{\textwidth}}
	\end{subfigure}  
	\begin{subfigure}[b]{0.195\textwidth}
		\includegraphics[trim={0.0cm 16cm 14cm 0cm},clip,width=\textwidth]{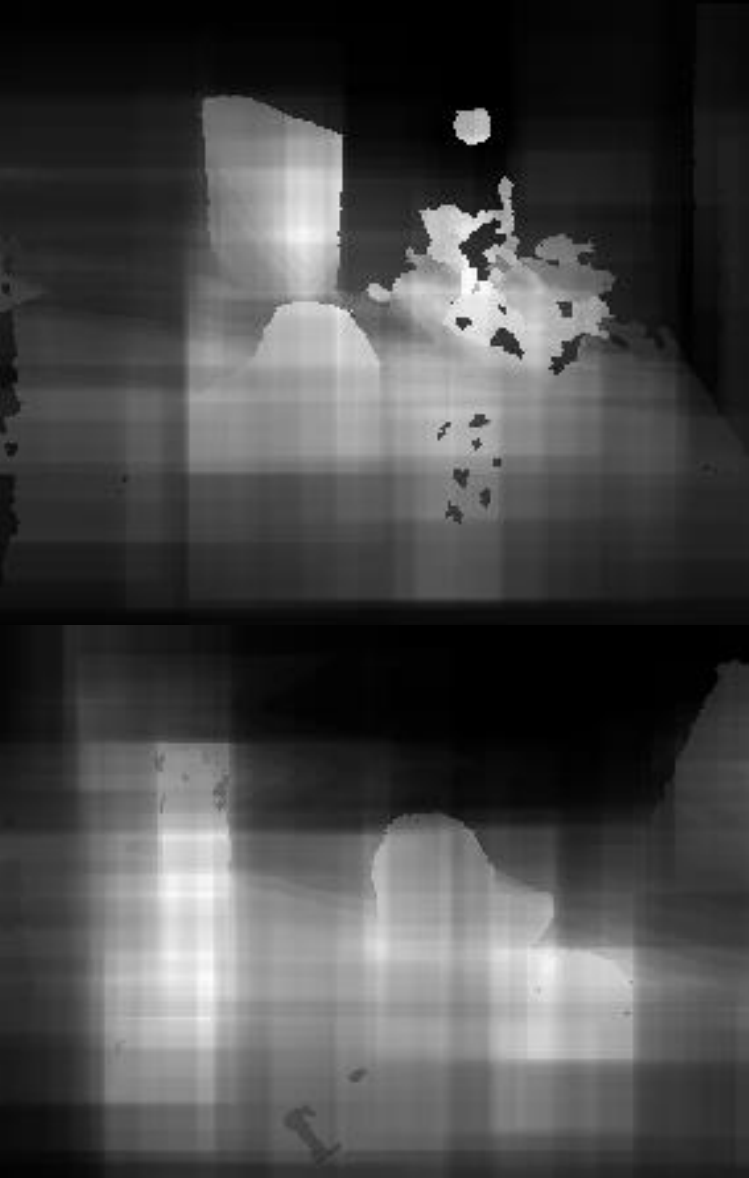}
		\caption{Weighted $2$D \\ \qquad heatmap $\bm{H}_{2D}$}
	\end{subfigure}  
	\begin{subfigure}[b]{0.195\textwidth}
		\includegraphics[trim={0.0cm 16cm 14cm 0cm},clip,width=\textwidth]{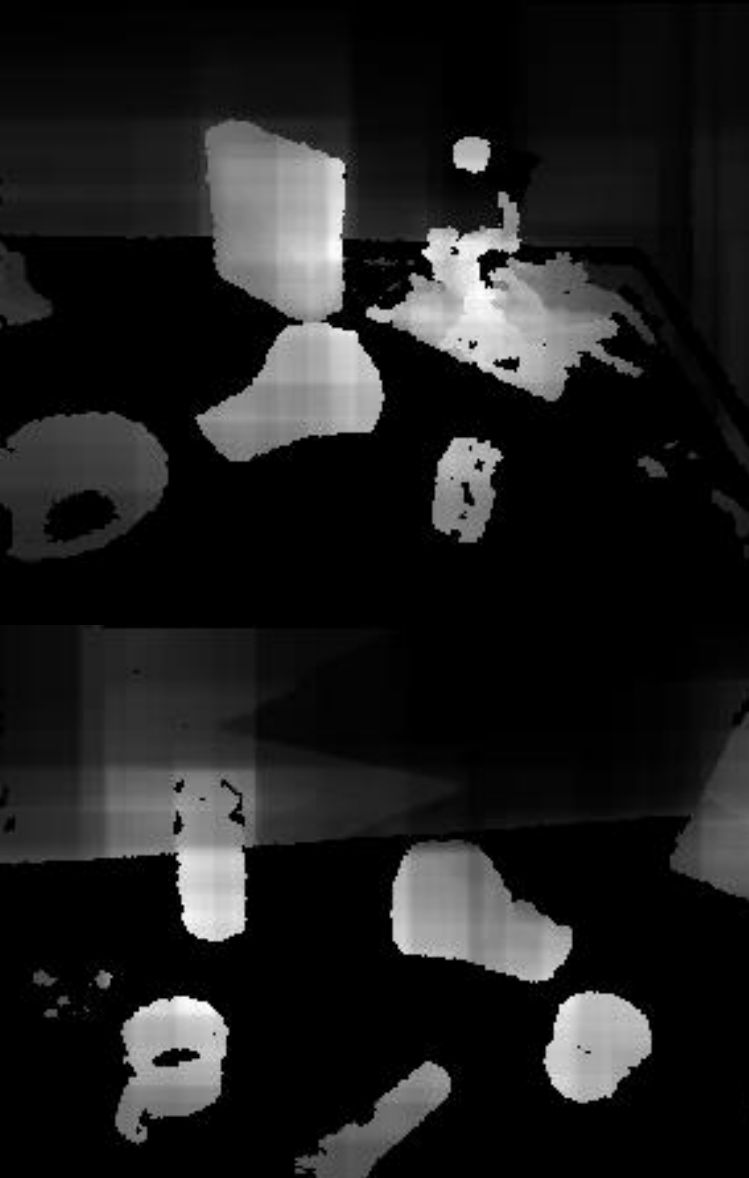}
		\caption{Heatmap after plane \\ \qquad removal $\tilde{\bm{H}}_{2D}$}
	\end{subfigure}  
		\begin{subfigure}[b]{0.195\textwidth}
		\includegraphics[trim={0.0cm 16cm 14cm 0cm},clip,width=\textwidth]{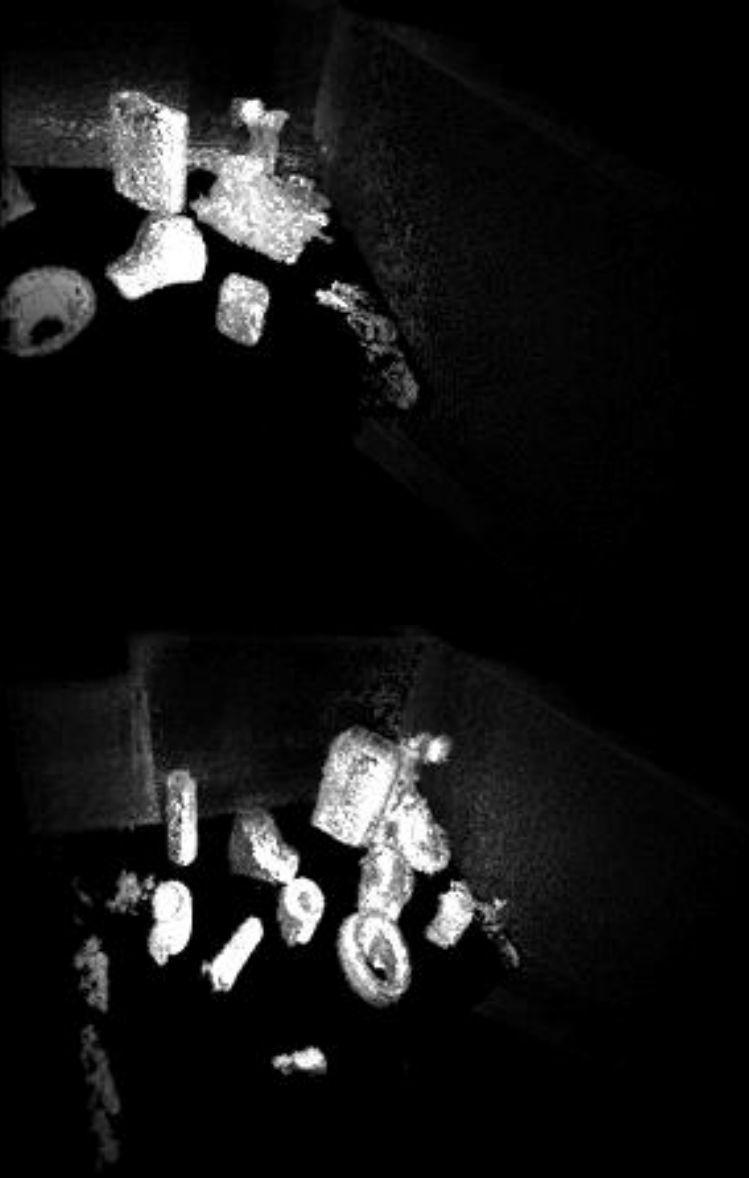}
		\caption{Global heatmap $\bm{H}_{3D}$  projected onto image plane}
	\end{subfigure}  
  \caption{Left column displays different RGB frames from \textit{table$\_1$} scene in \cite{lai_icra2011}. $2^{nd}$ column shows the matched pixels of the current frame with the previous frame. $3^{rd}$ column highlights our weighted $2$D heatmap after depth based filtering. $4^{th}$ column displays our refined heatmap after plane removal. Finally, the $5^{th}$ column shows the current global heatmap of the entire scene. The global heatmap accumulates the confidence of each point in the scene as the camera moves around in the scene.}
 \label{fig::global_heatmap}
\end{figure*}

Since we are only interested in removing the supporting plane underneath objects of interest, the obtained $2$D heatmap  is utilized to our advantage. Based on the inlier criterion defined in Eq.~\ref{eq::plane_eq}, we estimate the plane of interest for the first image frame by selecting the plane ${\bm{p}^j}^*$ that has the highest accumulated heat value as follows, 
\begin{equation}  
j^* =  \argmax_{j} \sum_{i|\bm{x}_i \in \bm{S}^j}\bm{H}_{2D}(\bm{x}_i)  \;.
\end{equation}
This heatmap based plane estimation assists us in consistently finding the correct plane of interest as shown in Fig.~\ref{fig::plane_removal}. For the next frame, the camera pose is used to compute the planar points by projecting onto the camera plane without needing to recompute the plane again. The pixels corresponding to the supporting plane are assigned zero confidence to obtain a filtered $2$D heatmap $\tilde{\bm{H}}_{2D}$: 
\begin{equation}\label{eq::final_2D_hetamap}
\tilde{\bm{H}}_{2D} [u,v] = \begin{cases} 
0 & \forall [u,v] \in \{\pi( \bm{K}, \bm{P}, \bm{x}_i) | \bm{x}_i \in \bm{S}^{j^*}\}  \\ 
\bm{H}_{2D}[u,v] & \text{otherwise} \;. \\ \end{cases}
\end{equation}
The plane parameters ${\bm{p}^j}^*$ are recomputed every ten frames to account for camera drift that SLAM methods often suffer from. The plane parameters are stored in a separate matrix ${\bm{P}}^*$ to be used later. Instead of repeating the entire plane estimation process, we utilize the information of knowing where the plane lies to compute the new plane of interest quickly and efficiently. We are able to remove the correct supporting plane using this automated plane removal approach consistently. One situation where this approach fails is where a camera trajectory starts with looking at the floor and slowly pans towards the table. The floor is selected as the plane of interest in initial frames when the table and other objects are not seen in camera's field of view. Once the table occupies enough regions in the camera's view, so that its accumulated confidence is higher than the floor, the table is selected from there-on as the plane of interest. In these intermediate frames the partially seen table is not removed and thus retains its high confidence value in the heatmap. We discuss how to resolve this issue in Sec.~\ref{sec:clustering}. We call the frames where we recompute plane parameters as {\it keyframes}. Unlike SLAM methods where keyframes are used to estimate the camera pose of the current frame, the keyframes are only used to project the plane onto the current frame, and assign the corresponding planar pixels zero confidence as shown in Fig.~\ref{fig::global_heatmap}(d).
\end{subsection}

\begin{subsection}{3D Heatmap Generation via Multi-view Fusion}\label{sec:registration_3D}

After the filtered $2$D heatmap $\tilde{\bm{H}}_{2D}$ is computed, the corresponding depth information available is used to project the points in $3$D. The next step is to fuse this information with the existing $3$D heatmap $\bm{H}_{3D}$.

A standard approach to fuse this information with the current $3$D heatmap is to find the closest points using techniques such as ICP \cite{rusinkiewicz2001}. However, such an approach quickly becomes computationally expensive, which deviates from our goal of developing a fast approach for $3$D object proposals. Another way to tackle this problem is to use the poses estimated by depth based SLAM methods to allocate points in $3$D voxels or Octrees. However, due to the sheer amount of $3$D points and span of the room, such an approach also requires a large amount of memory and incurs heavy computation loads. This issue is resolved by using image warping and creating a $2$D index table. First, the mapping from the $2$D heatmap to a $3$D heatmap is initialized for the first frame. This bijective mapping, called \textit{indexMap}, stores the location of each pixel of the current $2$D heatmap in the global $\bm{H}_{3D}$ heatmap.

Using Eq.~\ref{eq::warp}, the previous frame's depth information  is utilized to warp the image onto the current $i^{th}$ frame. We round the projected pixel location to the nearest integer and compare their depth and color values per pixel. Let us assume that pixel $\tilde{\bm{x}}_c$ of the previous frame is warped onto $\bm{x}_c$ of the current frame.
\begin{equation} \label{eq::warp}
\bm{x}_c =   \pi( \bm{K}, \bm{P}^i, \pi^{-1}( \bm{K}, \bm{P}^{(i-1)}, \tilde{\bm{x}}_c, \bm{Z}^{(i-1)}[\tilde{\bm{x}}_c] )
\end{equation}
Based on this information, we compute the difference in intensity and depth at each matched pixel:
\begin{equation}\label{eq:diff_I}
\Delta \bm{I}^i[\bm{x}_c] = ||\bm{I}^i[\bm{x}_c] - \bm{I}^{(i-1)}[\tilde{\bm{x}}_c]||_2
\end{equation} 
\begin{equation}\label{eq:diff_Z}
\Delta \bm{Z}^i(\bm{x}_c) = |\bm{Z}^i[\bm{x}_c] - \tilde{\bm{Z}}^{(i-1)}[\tilde{\bm{x}}_c]| \;,
\end{equation} 
where $\tilde{\bm{Z}}^{(i-1)}[\tilde{\bm{x}}_c]$ denotes warped depth value in the current frame. If projected pixel's color and depth information is within a threshold ($\epsilon_I$, $\epsilon_Z$) of the current pixel's information, the two corresponding pixels are considered a match and the index is copied to the current pixel. In case where more than one pixel from previous frame matches a pixel in the current frame, then the pixel corresponding to a lower warped depth value (foreground) is chosen as the matching pixel. This matching is shown in Fig.~\ref{fig::global_heatmap}(b). If no pixel from the previous frame matches the current pixel, the pixel is identified as a new point and added to the global $3$D heatmap along with the current \textit{indexMap}. Since this approach requires depth information at each pixel, if the depth information is not available or contains noisy depth, this can lead to wrongly matched pixels. We choose to ignore these pixels as our primary goal is obtaining fast and efficient $3$D proposals, and performing this step for every image can incur more computational load. Nevertheless, if more robust results are required, one can fill the holes using a nearest-neighbor method or an adaptive filter using the corresponding color images \cite{li_eccv16}.

For all matched points, the original $3$D point's confidence value is increased by the current matched pixel's confidence value and the counter is incremented by a unit. In addition, the color and location of the $3$D point is adjusted by a weighted average of the current $3$D location of the pixel and the global location of the matched point. This accounts for any minor drift error that might occur while estimating the camera trajectory in an indoor environment.

\end{subsection}

\begin{subsection}{3D Heatmap Filtering using Average Confidence Measure}\label{sec:point_selection} 

Once our weighted $3$D heatmap is obtained, any points that are seen less than five times or $5\%$ of the total number of frames, whichever is lower, are identified as unwanted points and discarded. We further use a metric: pseudo-average confidence to rank the global $3$D points. The pseudo-average confidence of the $i^{th}$ point is computed as:
\begin{equation}
\bar{c}(i) = \frac{\bm{H}_{3D}(i; c)}{\bm{H}_{3D}(i; f) + \tau}, \qquad i = 1, \hdots, N \;,
\end{equation}
where $\bm{H}_{3D}(i; c)$ denotes retrieving the $c$ (confidence) element of the $i^{th}$ point stored in the 3D heatmap $\bm{H}_{3D}$, similarly for $\bm{H}_{3D}(i; f)$ that returns the $f$ (frequency) element. $\tau$ is a constant introduced to gently favor points seen more often.

The pseudo-average confidence ranks points seen more often higher than points seen less often. Intuitively, this makes sense as we should see the $3$D points lying on objects of interest more often than other points. 
Only those points are retained that have $\bar{c}(i) \geq {\epsilon}$. This ensures that we obtain a good precision while maintaining an acceptable recall value. If a better precision or recall is desired, one can increase or decrease this threshold respectively. An example of the top ranked $3$D point cloud is shown in Fig.~\ref{fig:ours_fig_algo}(a). 
\end{subsection}

\begin{figure*}[t!]
\centering
\includegraphics[trim={0.0cm 12.5cm 0.5cm 0.0cm},clip,width=0.95\textwidth]{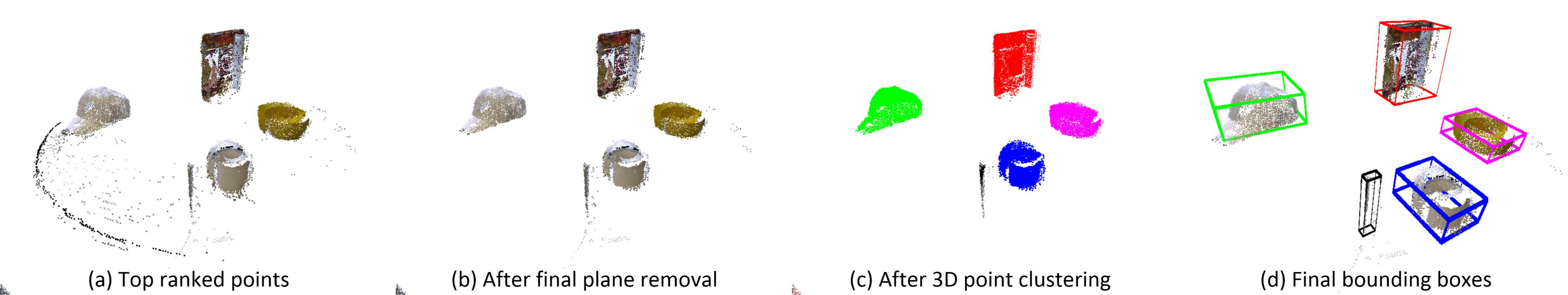} 
\caption{Once we obtain the top ranked points, we perform a final plane removal step to remove any leftover supporting plane points. We perform density-based spatial clustering to find the most dominant point clusters and estimate their bounding boxes.}
\label{fig:ours_fig_algo}
\end{figure*}

\begin{subsection}{3D Point Clustering and 3D Bounding Box Generation}
\label{sec:clustering} 

As discussed in Sec.~\ref{sec:plane_removal}, it is possible that the scene capture starts without the objects and table in the camera's field of view. The scene capture may start with room floor or walls in its entire field of view and then move to the area of interest. In cases where the table is only partially seen (a few pixels) when the camera field of view is moving towards the objects of interest, the accumulated heatmap value of the planes corresponding to walls and floors may be temporarily higher in a few image frames. In such cases, the floor and walls will be removed from heatmap but the table will be left untouched. Thus, some outer regions of the table may still be present in our final ranked points as seen in Fig.~\ref{fig:ours_fig_algo}(a). To remove these unwanted points, the plane parameters stored in the matrix ${\bm{P}}^*$ obtained from our previous keyframes are used. The plane parameters for each keyframe may correspond to the floor or vertical wall when the table is not seen. Thus, we first identify different planes that were removed using K-means (walls, rooms, table etc.) For each unique plane, the best fitting plane is estimated using these different plane parameters. Thereafter, a final plane removal step is performed on our top ranked points by finding any points that satisfy Eq.~\ref{eq::plane_eq} for the estimated plane. The final filtered $3$D point cloud is shown in Fig.~\ref{fig:ours_fig_algo}(b).

After plane removal, we perform density based clustering using density-based spatial clustering of applications with noise (DBSCAN) \cite{DBScan} is performed on our filtered top ranked points. DBSCAN groups together points that are closely packed and marks points as outliers that lie alone in low-density regions. This allows us to reject any points that may belong to uninteresting regions such as walls and table that may exist in our top filtered points. The advantage of using a density based approach over other techniques such as K-means is that we do not need to specify the number of clusters that are present in our filtered data. Thus, depending on the scene complexity, this approach can automatically select the relevant number of regions of interest in the scene. DBSCAN can run in an overall average runtime complexity of $O(n log (n))$. An example of DBSCAN based clustering is shown in Fig.~\ref{fig:ours_fig_algo}(c).

After density based clustering, a tight $3$D bounding box is estimated for each cluster. However, finding a good bounding box is challenging in the camera's frame of reference, since the horizontal direction (parallel to the floor) may not be the same as the camera's horizontal direction. First a normal to the supporting horizontal plane, $\bm{n}_c$, is estimated by using the entire $3$D point cloud obtained after our data collection as explained in Sec.~\ref{sec:plane_removal}. Thereafter, each cluster is transformed to the world coordinate frame where the orthogonal directions $X$, $Y$ and $Z$ match the standard normal vectors $[1,~0,~0]^\intercal$, $[0,~1,~0]^\intercal$ and $[0,~0,~1]^\intercal$ respectively using Eq.~\ref{eq::rotation_BB}. Thereafter, we find the minima and maxima in each of the three orthogonal directions and draw a bounding box around the clusters. This bounding box is then transformed back to the original coordinate frame.
\begin{eqnarray}\label{eq::rotation_BB}
\bm{v} & = &  \bm{n}_c \times \bm{n}_w, \qquad 
s = ||\bm{v}||_2, \qquad 
c = \bm{n}_c \cdot \bm{n}_w, \nonumber \\
\bm{R} & = & \bm{I} + [\bm{v}]_\times + \frac{(1-c)}{s^2}[\bm{v}]_{\times}^2   
\end{eqnarray}
where $\bm{n}_c$ denotes the normal to the detected horizontal supporting plane in the camera's coordinate frame, and $\bm{n}_w = [0,1,0]^\intercal$ is the world coordinate's normal direction. $[\bm{v}]_\times $ is the skew symmetric matrix of the vector $\bm{v}$. If the bounding boxes of neighboring clusters intersect, we combine those clusters together. This usually happens when an object breaks into two sub-parts due to pose estimation errors or in the presence of specular objects such as a soda can. Any bounding boxes that have a volume of less than $1cm^3$ are rejected as these small patches are usually a part of the walls or floor plane. The final refined $3$D point cloud with the respective bounding boxes are shown in Fig.~\ref{fig:ours_fig_algo}(d). This modular approach assists in finding tight bounding boxes in the direction of gravity for each object of interest.

\end{subsection}
 
\end{section}

\begin{section}{Experimental Evaluation}\label{sec:results}

We use four datasets to conduct our experimental analysis and evaluation: object segmentation dataset \cite{OSD_dataset}, UW-RGBD scene dataset \cite{lai_icra14}, RGBD scene dataset 2011 \cite{lai_icra2011}, and our own dataset.

The object segmentation dataset (OSD) \cite{OSD_dataset} consists of $111$ labeled RGBD images in six subsets. The images vary from having two small objects on a table to containing more than ten objects stacked side by side and on top of each other. The dataset provides pixelwise labeling of each instance of the objects. We use these labeled images to create $2$D ground-truth bounding boxes for each object present in the dataset to enable learning and for precision-recall evaluation. The depth images are pre-aligned with the color images. We primarily used this dataset to train our parameters.

The UW-RGBD scene dataset \cite{lai_icra14} contains $14$ scenes reconstructed from RGB-D video sequences containing furniture and some table-top objects such as caps, cereal boxes and coffee mugs. The scenes contain depth and color frames from a video collected by moving around the scene.  The dataset provides a globally labeled $3$D point cloud. We used Dense Visual SLAM \cite{kerl2013dense} to obtain the camera pose per frame to fuse the frames together. Our $3$D point cloud is slightly misaligned with the ground-truth point cloud by a few $mm$. It is not trivial to align these point clouds due to different error characteristics of the point clouds. Thus, for our current evaluation, we decided to ignore this mismatch as the misalignment is marginal. We primarily use this dataset for our $3$D evaluation and comparison with existing state of the art techniques in $2$D by projecting our bounding boxes on the image plane.

The RGBD scene dataset 2011 \cite{lai_icra2011} consists of eight realistic environment scenes of a lab, kitchens and office rooms. The objects of interest are placed on kitchen slabs and tables in a room. The dataset provides ground-truth $2$D bounding boxes for various objects of interest such as soda cans, caps, and flashlight. However, some objects such as laptops, computer mice and other kitchen equipment are labeled as background in this dataset. The authors used depth and color images to segment and identify the objects present in each frame independently. We use this dataset to show our results in a more cluttered environment, and compare our results in $2$D by projecting our bounding boxes on the image plane and comparing with the $2$D ground-truth bounding boxes.

The existing RGBD datasets are intended for benchmarking category-independent object segmentation and identification purposes, and thus only provide limited test cases. The effect of errors in SLAM algorithms is also ignored. Therefore, we collected our own sequence with more challenging situations where the scene is crowded and the objects are placed in proximity. We also take into consideration of imperfect SLAM and demonstrate that our method is only marginally affected given inaccurate camera poses estimation. We plan to make our dataset public in future.

\begin{subsection}{Evaluation of Single-frame 2D Object Proposals}
{

\textbf{OSD.} We first evaluate results on the object segmentation dataset (OSD) \cite{OSD_dataset} to demonstrate how our depth based filtering improves edge-boxes~\cite{zitnick2014edge}. We use this dataset to train our parameter $\epsilon_{\Delta}$ in Eq.~\ref{eq::bg_fg_classification}. We used $1000$ object proposals per image on the training subset and found $\epsilon_{\Delta} = 0.5m$ to give us the best results on the test dataset as well. Using depth based $2$D object proposal filtering as described in Sec.~\ref{sec:2D_heatmap}, we are successful in rejecting $5.1\%$ of the object proposals provided by edge-boxes across $111$ images.

From here on, the different parameters discussed in Sec.~\ref{sec:2D_heatmap} and \ref{sec:3D_fusion} are fixed as : $\{\epsilon_{\Delta} \text{, } \epsilon_{min}\text{, } \epsilon_{max} \text{, } \tau \} =  [0.5m \text{, }2cm \text{, }1m \text{, }10]$ and $\{\epsilon_{p} \text{, } \epsilon_{I} \text{, } \epsilon_{Z} \text{, }\epsilon\} = [0.005 \text{, } 0.05 \text{, } 0.01\text{, } 0.25]$.

\textbf{UW-RGBD scene dataset.} We also computed the acceptance and rejection rate on the UW-RGBD dataset for our depth based filtering process. Based on the background masking and culling odd-sized proposals as described in Sec.~\ref{sec:soft_filtering} and \ref{sec:hard_filtering},  $7\%$ object proposals are rejected, $31.3\%$ are fully accepted and $61.7\%$ object proposals undergo partial filtering to mask the background on the entire dataset.

\textbf{RGBD scene dataset 2011.} We use the ground-truth $2$D bounding boxes provided in the RGBD scenes dataset to report the average precision, recall and success rate. As here we treat and evaluate each frame independently, we do not perform multi-view fusion as described in Sec.~\ref{sec:3D_fusion}. Instead, after depth based filtering, we remove the unwanted supporting plane in $3$D (Sec.~\ref{sec:plane_removal}) and cluster the $3$D points to compute $3$D bounding boxes (Sec.~\ref{sec:clustering}). We project the $3$D points inside each of the $3$D bounding boxes back to the image plane, and compute the $2$D bounding boxes around these pixels. As noted earlier, most $2$D object proposal techniques aim for a high recall. In contrast, our goal is fast precise $3$D object proposals, we typically obtain $5$-$20$ object proposals depending on the scene complexity.

Let $BB_g(i)$ be the $i^{th}$ ground-truth $2$D object proposal, $BB_e(j)$ and $BB_o(j)$ be the $j^{th}$ edge-boxes and our $2$D object proposals, respectively. Let $M_e$ and $M_o$ be the total number of object proposals computed by edge-boxes and our method. First, we use the standard definition of Intersection over Union (IoU):
\begin{equation}\label{eq::IoU_ori}
\text{IoU}(i) = \max_{j}\left(\frac{\text{BB}_\text{g}(i) \cap \text{BB}_\text{o}(j)}{\text{BB}_\text{g}(i)  \cup \text{BB}_\text{o}(j)}\right) \quad \substack{\forall j \text{ } \in  \text{ } 1, \hdots , M \\  \\  \forall i  \text{ } \in \text{ } 1, \hdots,  N}
\end{equation}
where $M$ and $N$ are the total number of output object proposals and ground-truth object proposals respectively in a given scene. Detection rate (DR), sometimes also referred to as average recall is defined as:
\begin{equation} \label{eq::DR}
DR  =  \frac{1}{N \cdot K}\sum_{k=1}^K\sum_{i=1}^{N} (\text{IoU}(i;k) \geq 0.5)  \;,
\end{equation}
where $K$ is the total number of scenes in the dataset and $\text{IoU}(i;k)$ refers to the IoU for the $i^{th}$ ground-truth object proposal in the $k^{th}$ scene. We define a modified Intersection over Union ($\text{IoU}_{o}$) for the $j^{th}$ object proposal as follows,
\begin{equation}\label{eq::IoU}
\text{IoU}_o(j) = \max_{i}\left(\frac{\text{BB}_\text{g}(i) \cap \text{BB}_\text{o}(j)}{\text{BB}_\text{g}(i)  \cup \text{BB}_\text{o}(j)}\right) \quad \substack{\forall j \text{ } \in  \text{ } 1, \hdots , M \\  \\  \forall i  \text{ } \in \text{ } 1, \hdots,  N}
\end{equation}
The difference between our modified IoU and standard IoU is that we estimate the best intersection per output object proposal, while the standard IoU is computed per ground-truth object proposal. Our definition heavily penalizes any redundant object proposals. We obtain a zero IoU score for object proposals that do not intersect with ground-truth object proposals. Based on this definition, similar to Eq.~\ref{eq::DR}, we define Success Rate ($SR$) as:
\begin{equation}
SR  =  \frac{1}{M \cdot K}\sum_{k=1}^K\sum_{j=1}^{M} (\text{IoU}_o(j;k) \geq 0.5) \;,
\end{equation}
where $K$ is the total number of scenes in the dataset and $\text{IoU}_o(j;k)$ refers to our modified IoU for the $j^{th}$ object proposal on the $k^{th}$ scene. Success rate can also be interpreted as signal to noise ratio (SNR) and has been used previously in \cite{Alexe_CVPR2010} for $2$D object proposal analysis. We use different numbers of input $2$D edge-boxes proposals varying from $50$ to $2,000$ and report our results in Table~\ref{table::EB_ours_comp}. 

\begin{table}[t]
\begin{center}
{ \begin{tabular}{ |c|c|c|c|c| }
\hline
{Num. of} & \multicolumn{2}{c|}{{Edge-boxes~\cite{zitnick2014edge}}} & \multicolumn{2}{c|}{Ours}  \\ 
\cline{2-5}
{proposals} & SR & DR & SR & DR \\ \hline
$50$ & $0.21$ & $0.72$ & $0.66$ & $0.80$  \\ \hline
$100$ & $0.16$ & $0.78$ & $0.67$ & $0.83$  \\ \hline
$500$ & $0.08$ & $0.88$ & $0.69$ & $0.86$  \\ \hline
$1000$ & $0.05$ & $0.89$ & $0.67$ & $0.86$  \\ \hline
$2000$ & $0.03$ & $0.90$ & $0.67$ & $0.86$   \\ \hline
\end{tabular}
}
\end{center}
 \caption{Comparison between edge-boxes~\cite{zitnick2014edge} and our technique per frame on the RGBD scene dataset 2011~\cite{lai_icra2011}. SR and DR refer to success rate and detection rate respectively. }
\label{table::EB_ours_comp}
\end{table}
\begin{figure*}[t]
\centering
	\begin{subfigure}[b]{0.19\textwidth}
	\includegraphics[width=\textwidth]{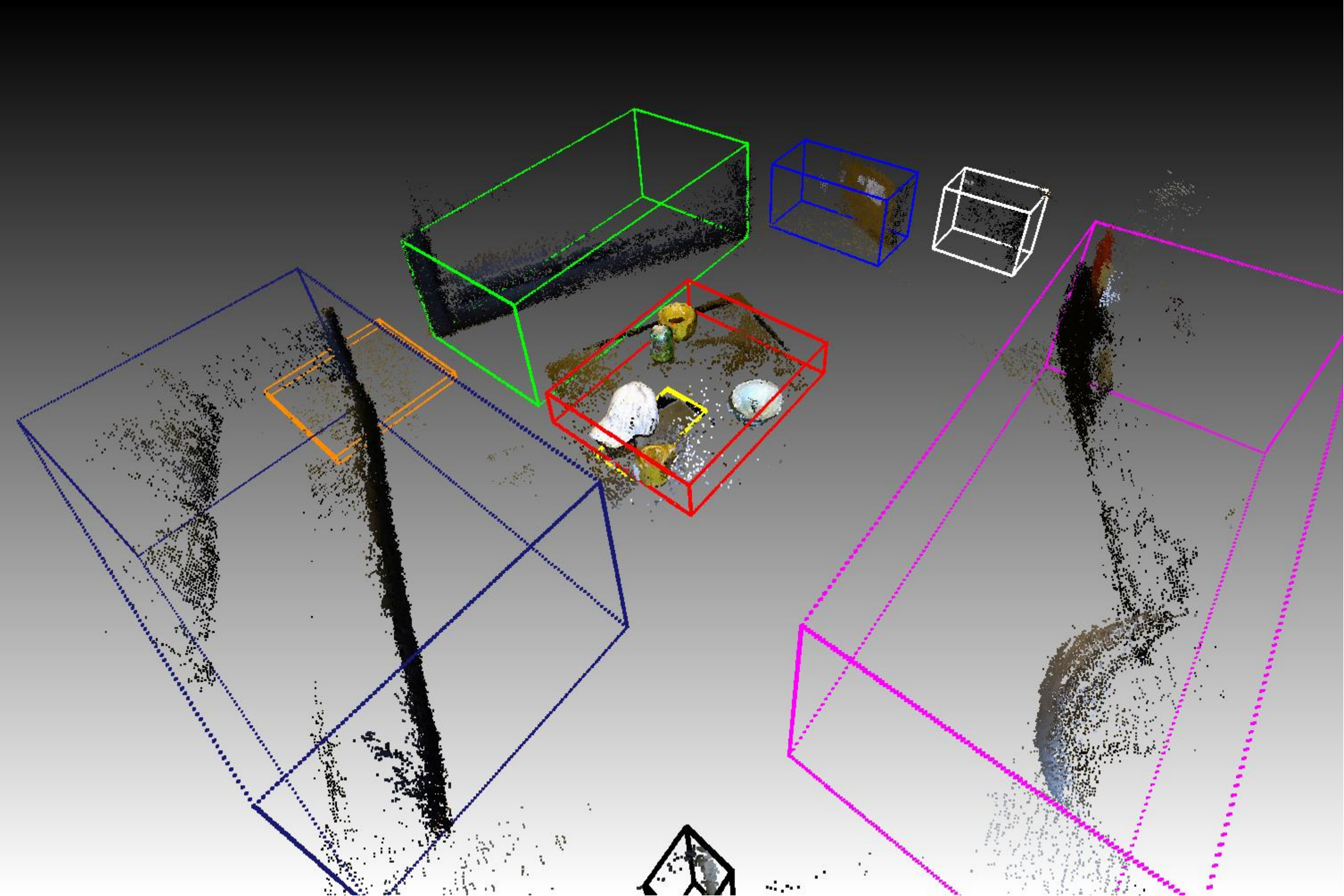}
	\caption{$noHmap$}
	\end{subfigure}
  \begin{subfigure}[b]{0.19\textwidth}
	\includegraphics[width=\textwidth]{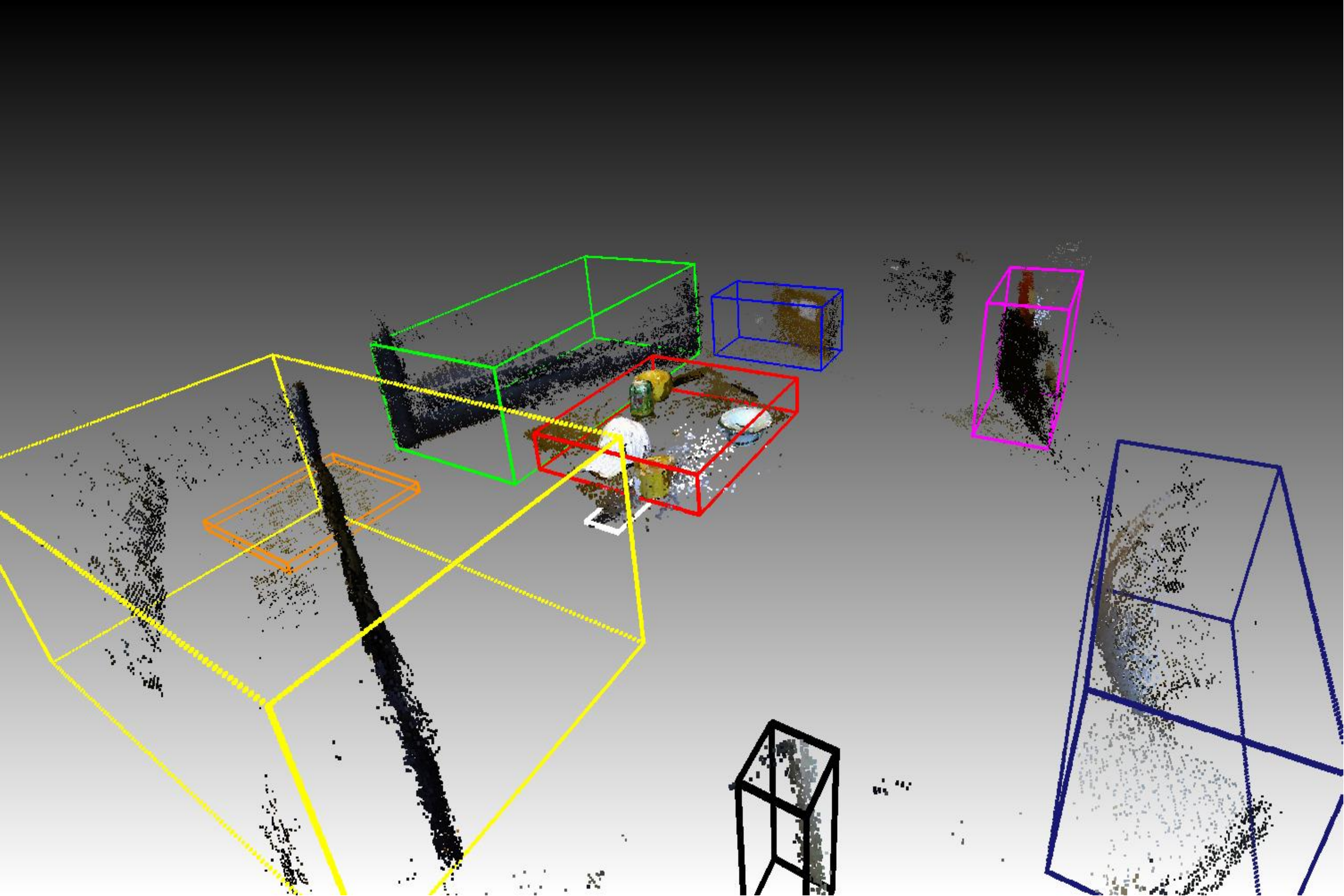}
	\caption{$noSFilt$}
	\end{subfigure}
  \begin{subfigure}[b]{0.19\textwidth}
	\includegraphics[width=\textwidth]{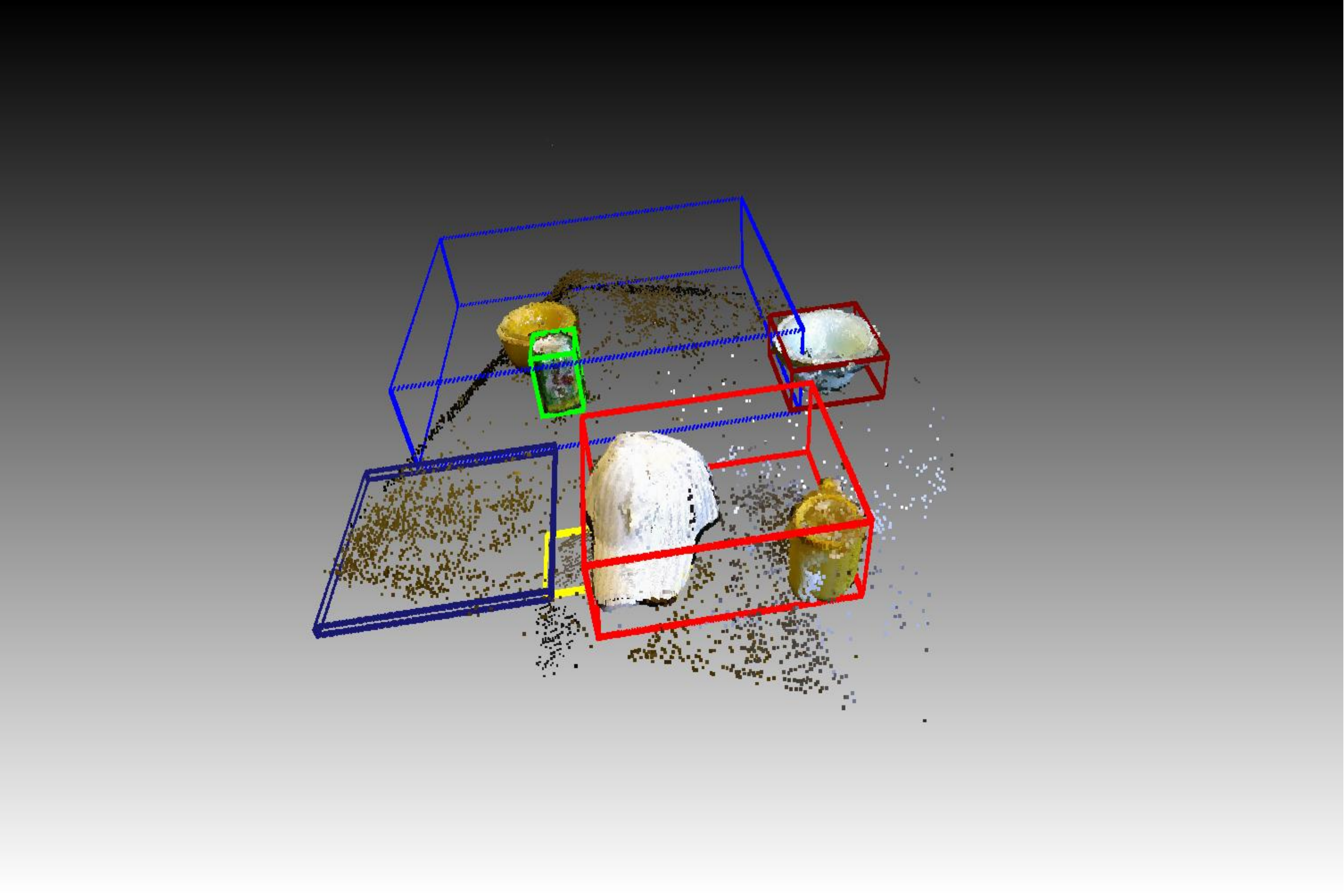}
	\caption{$noHFilt$}
	\end{subfigure}
  \begin{subfigure}[b]{0.19\textwidth}
	\includegraphics[width=\textwidth]{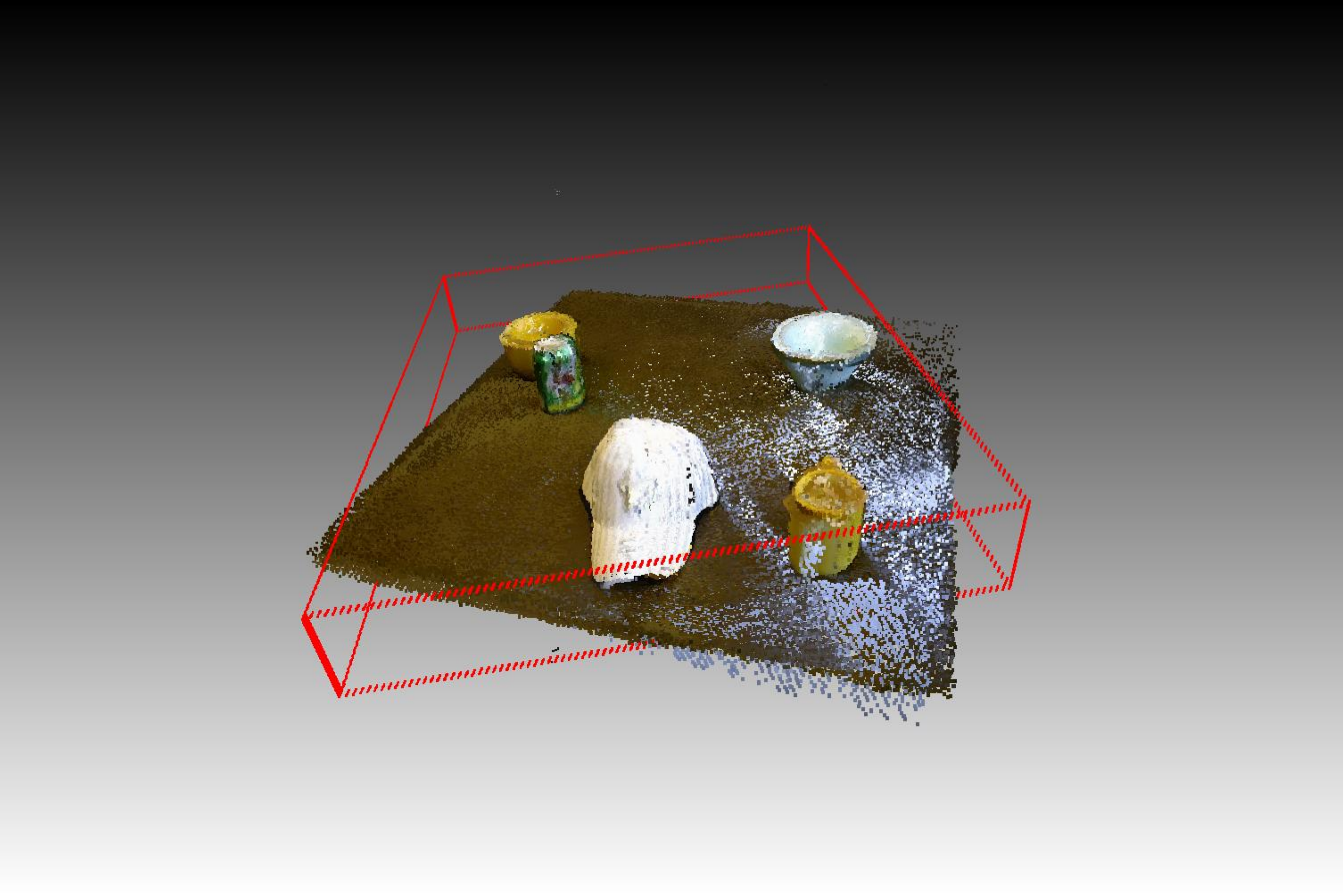}
	\caption{$noPlRem$}
	\end{subfigure}
	\begin{subfigure}[b]{0.19\textwidth}
	\includegraphics[width=\textwidth]{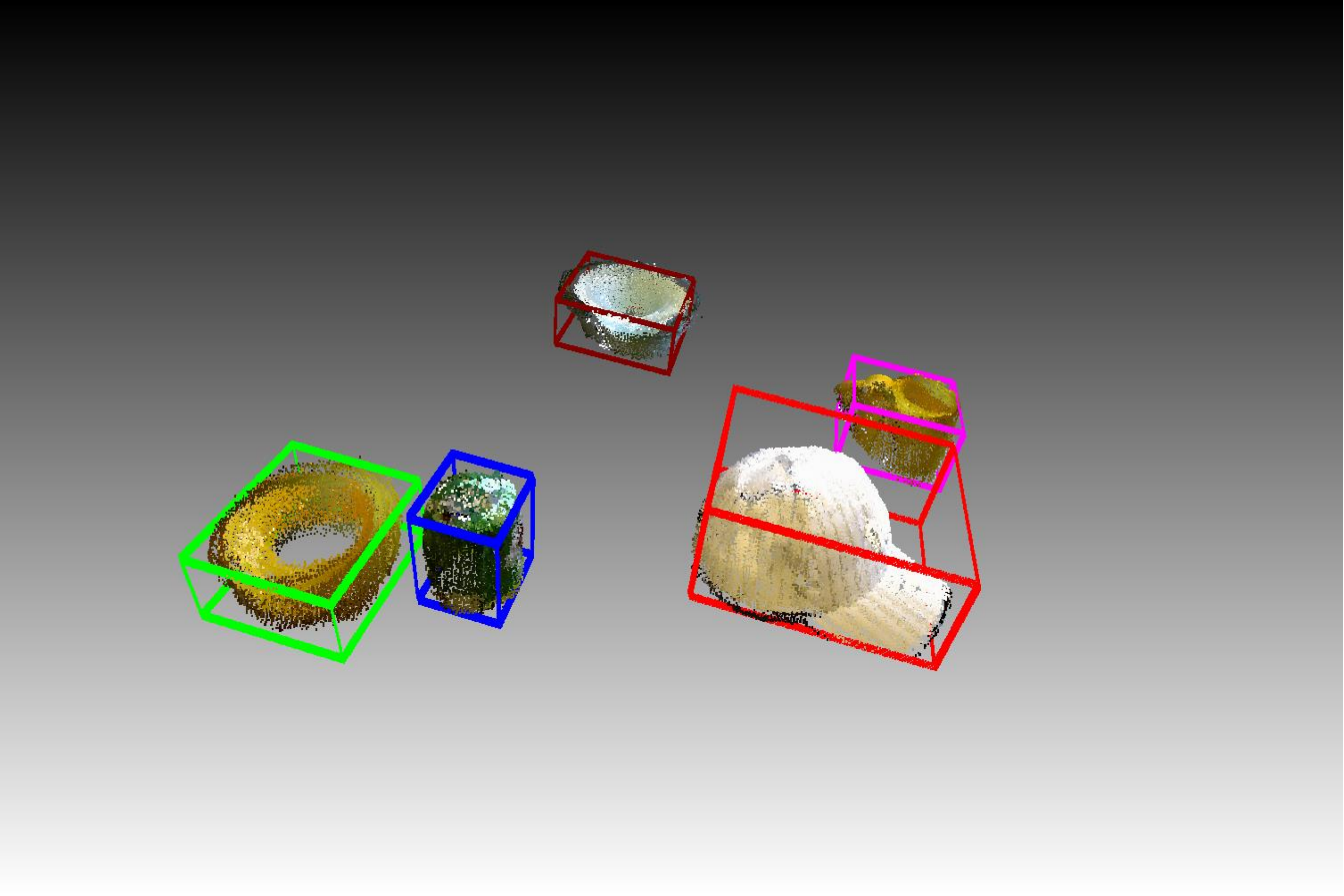}
	\caption{$Ours$}
	\end{subfigure}
  \caption{We show final results for a sample scene in the UW-RGBD scene dataset for our approach after removing one component. We obtain results in (a) without depth based filtering as described in Sec.~\ref{sec:2D_heatmap}. Results in (b) and (c) are obtained when background masking and bounding box rejection are not performed, respectively. (d) shows results without plane removal. (e) shows our result when all of these components (background masking, bounding box rejection, and plane removal) are used.}
 \label{fig::one_step_less}
\end{figure*}

On average we obtained $6.57$ output $2$D object proposals for our method. As shown in Table~\ref{table::EB_ours_comp}, using scene geometry enables us to significantly improve edge-boxes~\cite{zitnick2014edge}. We show remarkable improvement in success rate (SR). Our success rate is stable regardless of the number of input object proposals while the success rate drops drastically for edge-boxes. This is because the object proposals are chosen in order of their confidence values: the more number of input object proposals, the less likely they are to contain an object of interest in them. Our detection rate (DR) is higher than edge-boxes when using less numbers of proposals. It is slightly less than edge-boxes when using $1,000$ or more proposals. This is due to occluded objects present in the scene. Since this analysis is frame independent, our method only looks at the current frame and finds tight bounding boxes around objects seen in the frame currently. With a higher number of input proposals, edge-boxes is able to propose $2$D bounding boxes for these fully or partially occluded objects which increases its detection rate. We are able to overcome this issue by leveraging on multi-view information, and also our ultimate goal is not per-view detection rate.

}
\end{subsection}

\begin{subsection}{Evaluation of 2D and 3D Object Proposals from Multi-view Frames}

\textbf{UW-RGBD scene dataset.} After we obtain a $3$D heatmap per frame, we use multi-view information to fuse this information together. 
We use the pose estimated by Dense Visual SLAM \cite{kerl2013dense}. It uses depth images along with color information to create a globally consistent $3$D map and outputs a camera pose per frame. However, like other real-time SLAM algorithms that do not perform expensive optimization of camera poses and $3$D point clouds, it also suffers from camera drift errors. Our algorithm is designed such that it is tolerant to small drift and noisy $3$D point clouds. We generate object proposals at the same time as SLAM, and do not require the whole sequence to be captured first. 

\begin{table}[t]
\begin{center}
{ \begin{tabular}{ |c|c|c| }
\hline
Method & VGA & $\downarrow 2 $  \\ \hline
Overall & $\textbf{3.03s}$ & $\textbf{0.973s}$  \\ \hline \hline
EdgeBoxes \cite{zitnick2014edge} & $0.251s$ & $0.078s$  \\ \hline
Depth based filtering  & $0.373s$ & $0.195s$ \\ \hline
Plane removal & $0.14s$ & $0.06s$   \\\hline
Allocating $3$D points: &  \multicolumn{2}{c|}{{}}    \\\hline
Confidence and frequency  & $0.506s$ &  $0.132s$   \\ \hline
Location and color  & $1.481s$ & $0.359s$ \\ \hline
\end{tabular}
 }
\end{center}
 \caption{Analysis of average run-time performance per frame of our approach. Our experiments were conducted on a single-core Intel Xeon $E5$-$1620$ CPU.}
\label{table::time_eval}
\end{table}

The entire process takes $3.03s$ for VGA resolutions on average in MATLAB. However, a majority of the time ($>2.0s$) is spent in storing and accessing the $3$D point cloud. A detailed time evaluation of various techniques used in our approach is reported in Table~\ref{table::time_eval}. As we aim for a fast and efficient algorithm that is capable of online processing of $3$D object proposals, we downsample the images by $2$. This reduces the time taken per frame to less than one second in MATLAB on a single-core CPU. We demonstrate our results after downsampling in Fig.~\ref{fig:downsampld_res}.

First, we show the average IoU, both in $2$D and $3$D, obtained per object of interest for the UW-RGBD scene dataset \cite{lai_icra14} to demonstrate the effectiveness of each of our contributions in Table~\ref{table::one_step_less}. We estimate the $2$D bounding boxes by projecting the $3$D points inside each bounding box onto the image plane and computing a $2$D bounding box around it. We repeat this procedure to estimate ground-truth $2$D bounding boxes as well by using the ground-truth labeled point cloud. The left column in Table~\ref{table::one_step_less} represents one step that was skipped to estimate the $3$D bounding boxes. $noHmap$ refers to our approach without using the weighted heatmap $\bm{H}_{2D}$. In this approach, we skipped both soft filtering and hard filtering as described in Sec.~\ref{sec:2D_heatmap}. $noSFilt$ computes the results without performing background masking using depth information, and $noHFilt$ refers to our approach without rejecting odd-sized bounding boxes that are mainly part of the background. $noPlRem$ computes our results without performing plane removal per frame as described in Sec.~\ref{sec:plane_removal}. Finally, $Ours$ refers to our method that integrates all the steps together to compute the $3$D object proposals. We observe that each step is vital in improving the accuracy of our algorithm. A good supporting plane estimation plays an important role in removing the table underneath the objects for tight bounding box estimation. We show the final top ranked points along with the estimated bounding boxes in Fig.~\ref{fig::one_step_less}. 
\begin{table}
\centering
\resizebox{1.0\columnwidth}{!}{ \begin{tabular}{ |c|c|c|c|c||c|c|c|c| }
\hline
{\textbf{Method}} & \multicolumn{4}{c||}{{$\textbf{2D}$}} & \multicolumn{4}{c|}{{$\textbf{3D}$}}  \\ 
\cline{2-9}
{\textbf{Used}} & $IoU$ & $IoU_o$ & SR & DR & $IoU$ & $IoU_o$ & SR & DR \\ \hline
$noHmap$ & $0.32$ & $0.19$ & $0.17$ & $0.30$  & $0.17$ & $0.07$ & $0.01$ & $0.19$ \\ \hline
$noSFilt$ & $0.51$ & $0.25$ & $0.29$ & $0.60$  & $0.33$ & $0.15$ & $0.13$ & $0.31$\\ \hline
$noHFilt$ & $0.77$ & $0.47$ & $0.60$ & $\textbf{1.00}$ & $0.53$ & $0.32$ & $0.35$ & $0.53$ \\ \hline
$noPlRem$ & $0.13$ & $0.10$ & $0.02$ & $0.02$ & $0.03$ & $0.02$ & $0.00$ & $0.00$ \\ \hline
$Ours$ & $\textbf{0.78}$ & $\textbf{0.50}$ & $\textbf{0.62}$ & $\textbf{1.00}$  & $\textbf{0.55}$ & $\textbf{0.34}$ & $\textbf{0.45}$ & $\textbf{0.66}$ \\ \hline
\end{tabular}
}
\caption{Comparison of $2$D and $3$D average IoU, modified IoU, success rate and detection rate for objects of interest present in the UW-RGBD dataset. We report the benefits of each of our contributions: background suppression, rejection of odd-sized bounding boxes, and supporting plane removal.}
\label{table::one_step_less}
\end{table}
We also compare our results with \cite{pillai2015monocular}. Our approach considers chairs as objects of interest as seen  in Fig.~\ref{fig:downsampld_res}(b). However, since \cite{pillai2015monocular} treat chairs as background objects, we also leave them out of our analysis for a fair comparison in Fig.~\ref{fig::Iou_results_ours}. Once we obtain the $3$D bounding boxes, we compute all the points that lie inside the current bounding box and project them onto the current image plane. Thereafter, we estimate the $2$D bounding box that surrounds these projected points. We repeat this procedure for the ground-truth bounding boxes and compute the standard IoU using Eq.~\ref{eq::IoU_ori}. Our $2$D proposals (with and without downsampling) consistently outperform \cite{pillai2015monocular}.  In fact, $3$D intersection is extremely sensitive to noise. A small misalignment of $12.6\%$ in each of three orthogonal directions reduces the $3$D IoU to below $0.5$. While our  recall rate is good for low $3$D IoU, it drops below $0.5$ quickly due to the small mismatch between our point cloud and the ground-truth point cloud.
 
Like \cite{lai_icra14}, we also use the ground-truth labeling per point to obtain precision-recall results for our $3$D object proposals. Let us consider the ground truth points  labeled as soda can, bowl, cap, chair as points of interest ($\hat{\bm{P}}$) and the background such as sofa, table, floor as redundant points ($\hat{\bm{N}}$). We overlay our $3$D bounding boxes on the ground-truth point cloud and check if a point of interest $i$ lies inside or outside our proposals. We consider the points of interest that lie inside our object proposals as True Positives ($\bm{TP}$). The redundant points that lie inside our object proposals False Positives ($\bm{FP}$):
\begin{align}
\bm{TP} & = \sum_{i=1}^{N_g}(\hat{\bm{P}}(i) \in \text{BB}_\text{o}(j) )  \quad \forall j \in 1\hdots M \\
\bm{FP} & = \sum_{i=1}^{N_g}(\hat{\bm{N}}(i) \in \text{BB}_\text{o}(j) )  \quad \forall j \in 1\hdots M \;, 
\end{align}
where $N_g$ refers to the number of ground-truth points in a given scene. $\bm{TP}$ and $\bm{FP}$ are used to compute the average precision (AP), average recall (AR) and F-measure.

\begin{figure}[t!]
\centering
\includegraphics[height = 3cm,width=8cm]{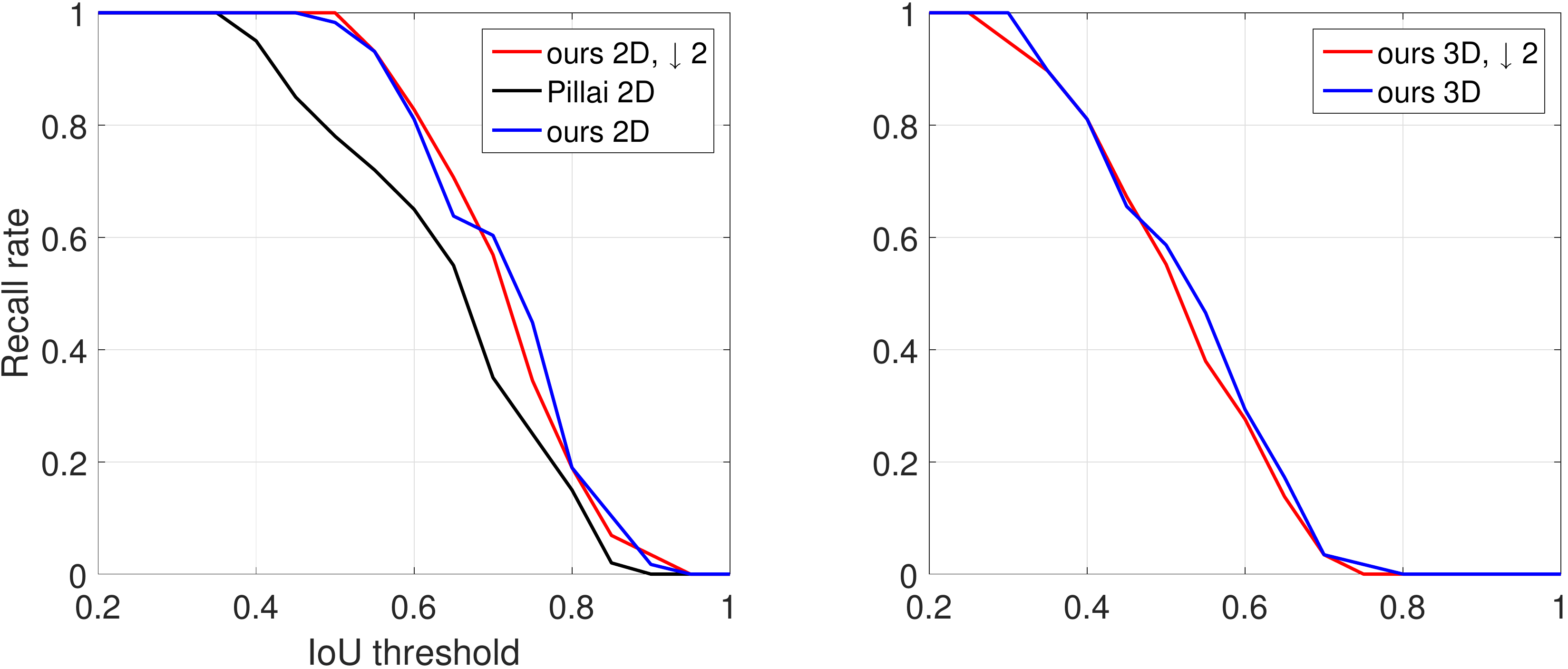}
\caption{Our $2$D and $3$D recall rate with varying threshold for IoU on the UW-RGBD dataset. Our $2$D proposals comfortably outperform \cite{pillai2015monocular}. As we exploit scene geometry, downsampling images by $2$ does not have an impact on our recall rate.}
\label{fig::Iou_results_ours}
\end{figure} 

We compare our object proposals with state-of-the-art segmentation and classification results reported on the UW-RGBD dataset in Table~\ref{table::time_comp}. Even for $3$D P-R measurements, we obtain the best mean AP, while maintaining an acceptable recall. The ground-truth labeling considers objects such as desktops as part of the background. This decreases our average precision as these detected objects are considered as false positives. If we ignore the scenes $13$ and $14$ in the dataset where we see these objects, our average precision increases from $93.46$ to $98.76$. Another issue affecting our recall performance is the presence of chairs. The camera's field-of-view is concentrated mainly on the objects on the scene, and hence the chair points (especially those under the seat) are not seen frequently and are filtered out by our technique. This reduces our recall rate if we consider chairs as objects of interest. We also compute our recall without chairs as objects of interest and are reported in Table~\ref{table::time_comp}. Our $3$D recall is lower than \cite{lai_icra14} because of the small misalignment between our point cloud and the ground-truth point cloud. This misalignment results in some points of interest ($\hat{\bm{P}}$) that are not included in our true positives ($\bm{TP}_k$), yielding an artificially lower recall rate.

Our technique is agnostic about the number of objects that may be present in the scene and hence, can scale well for crowded scenes that contain lots of objects. We obtain on average $6.57$ $3$D object proposals per scene. Our results can be further improved with an improved camera pose estimation, as in some cases the objects break into two or more discontinuous point clusters due to noisy camera poses. This results in multiple distinct object proposals for one object, essentially dividing the object into two or more pieces.

\begin{figure*}[t!]
\begin{subfigure}[b]{.16\textwidth}
\includegraphics[width=\textwidth]{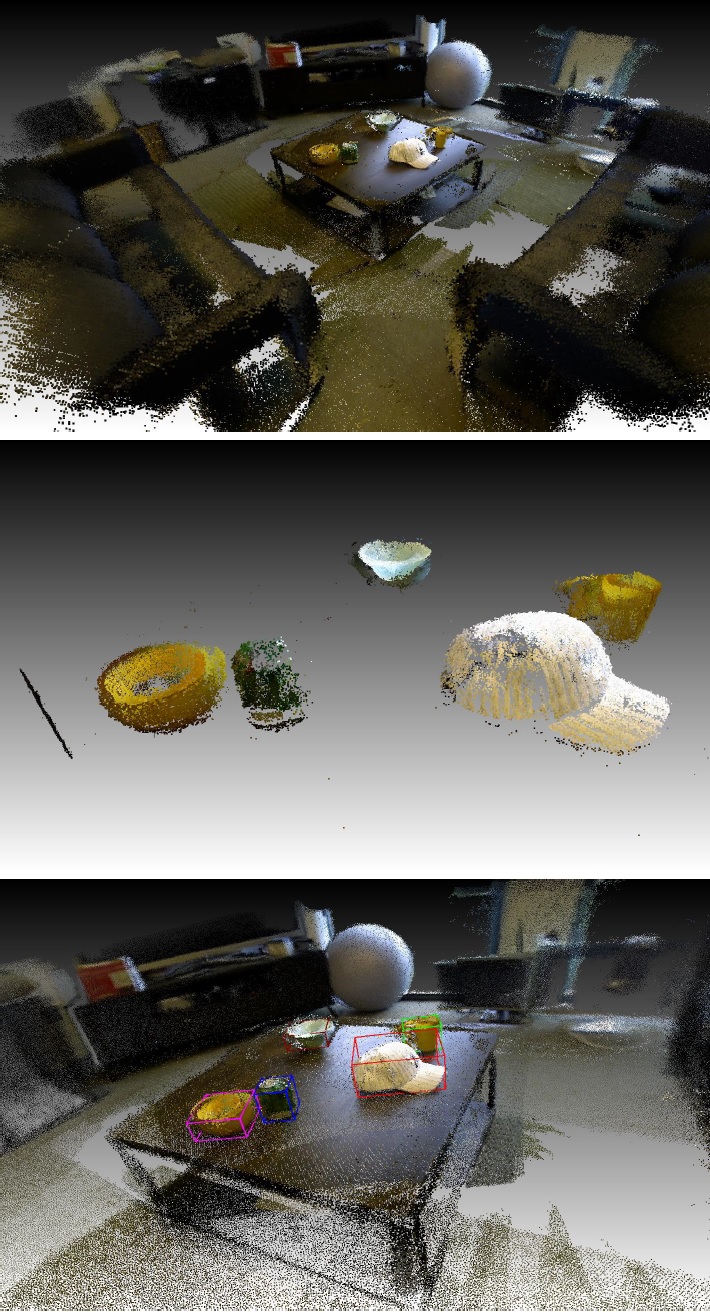} 
\caption{}
\end{subfigure} 
\begin{subfigure}[b]{.16\textwidth}
\includegraphics[width=\textwidth]{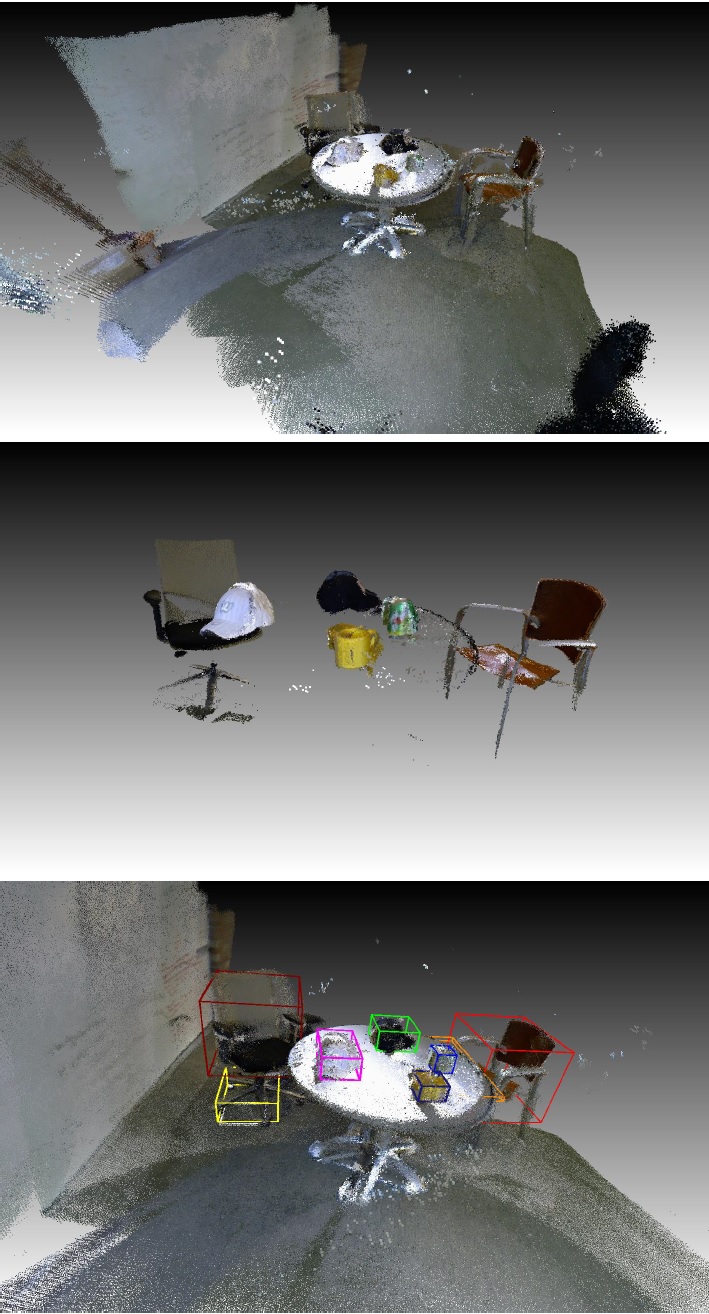} 
\caption{}
\end{subfigure} 
\begin{subfigure}[b]{.16\textwidth}
\includegraphics[width=\textwidth]{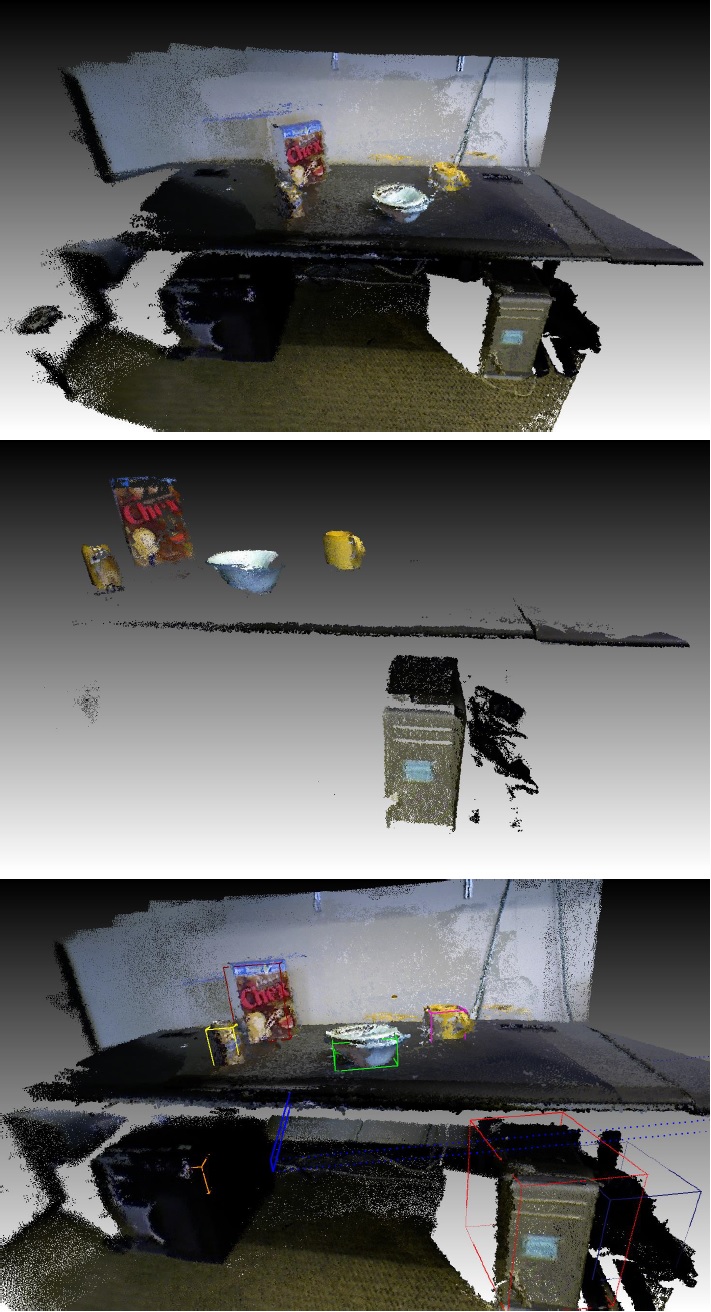} 
\caption{}
\end{subfigure} 
\rulesep
\begin{subfigure}[b]{.16\textwidth}
\includegraphics[trim={0.0cm 6cm 0.5cm 0.0cm},clip,width=\textwidth]{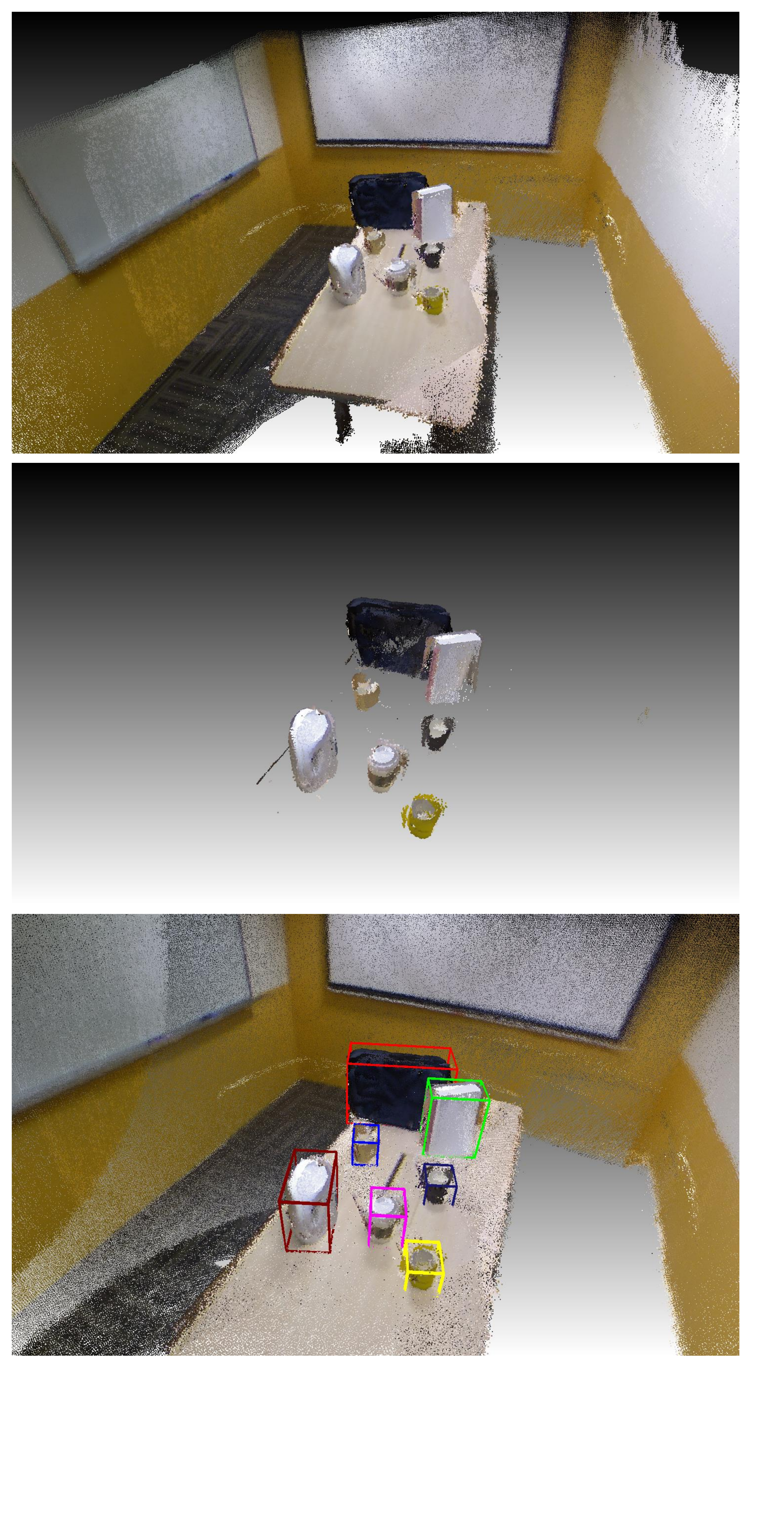} 
\caption{}
\end{subfigure} 
\begin{subfigure}[b]{.16\textwidth}
\includegraphics[trim={0.0cm 6cm 0.5cm 0.0cm},clip,width=\textwidth]{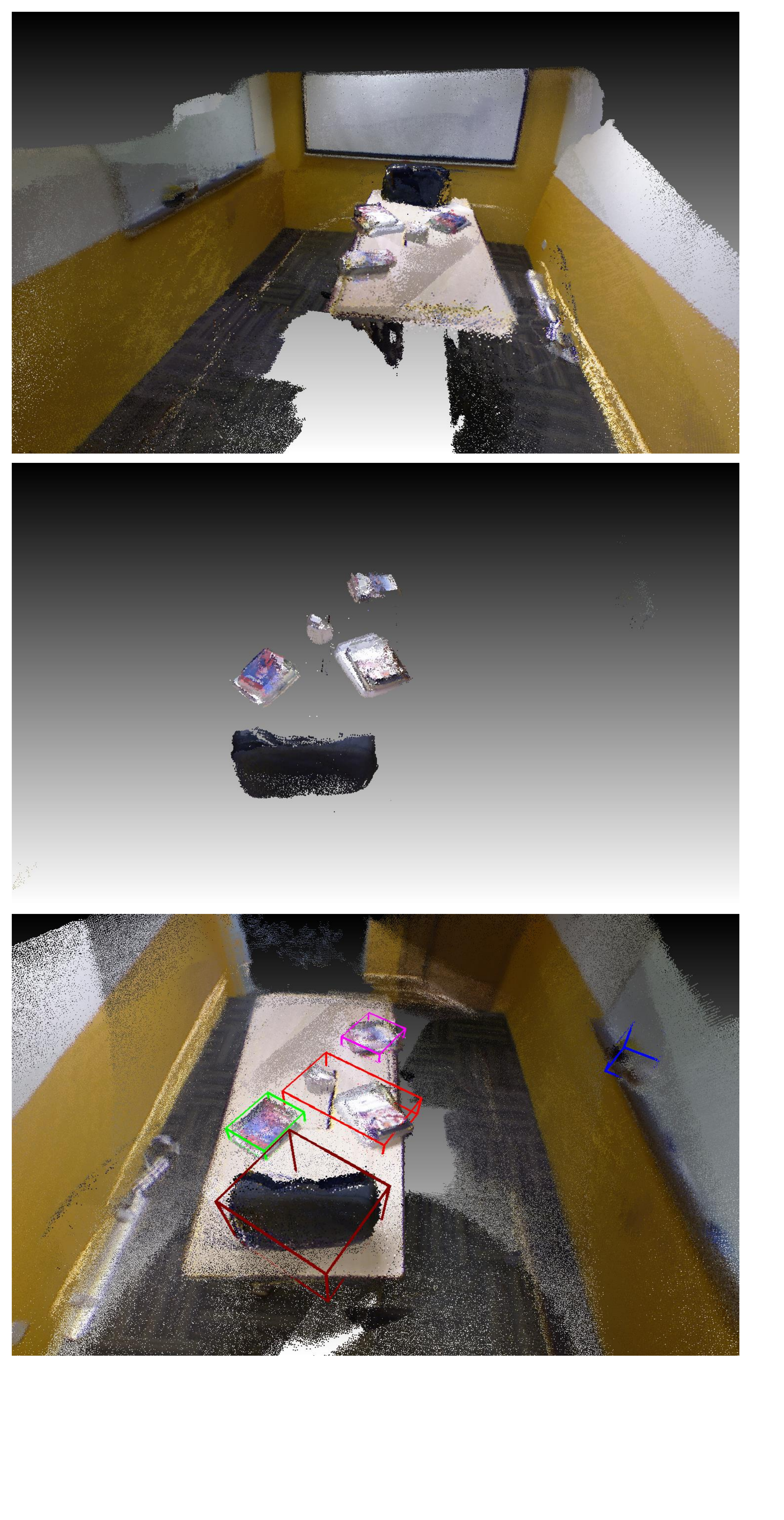} 
\caption{}
\end{subfigure} 
\begin{subfigure}[b]{.16\textwidth}
\includegraphics[trim={0.0cm 6cm 0.5cm 0.0cm},clip,width=\textwidth]{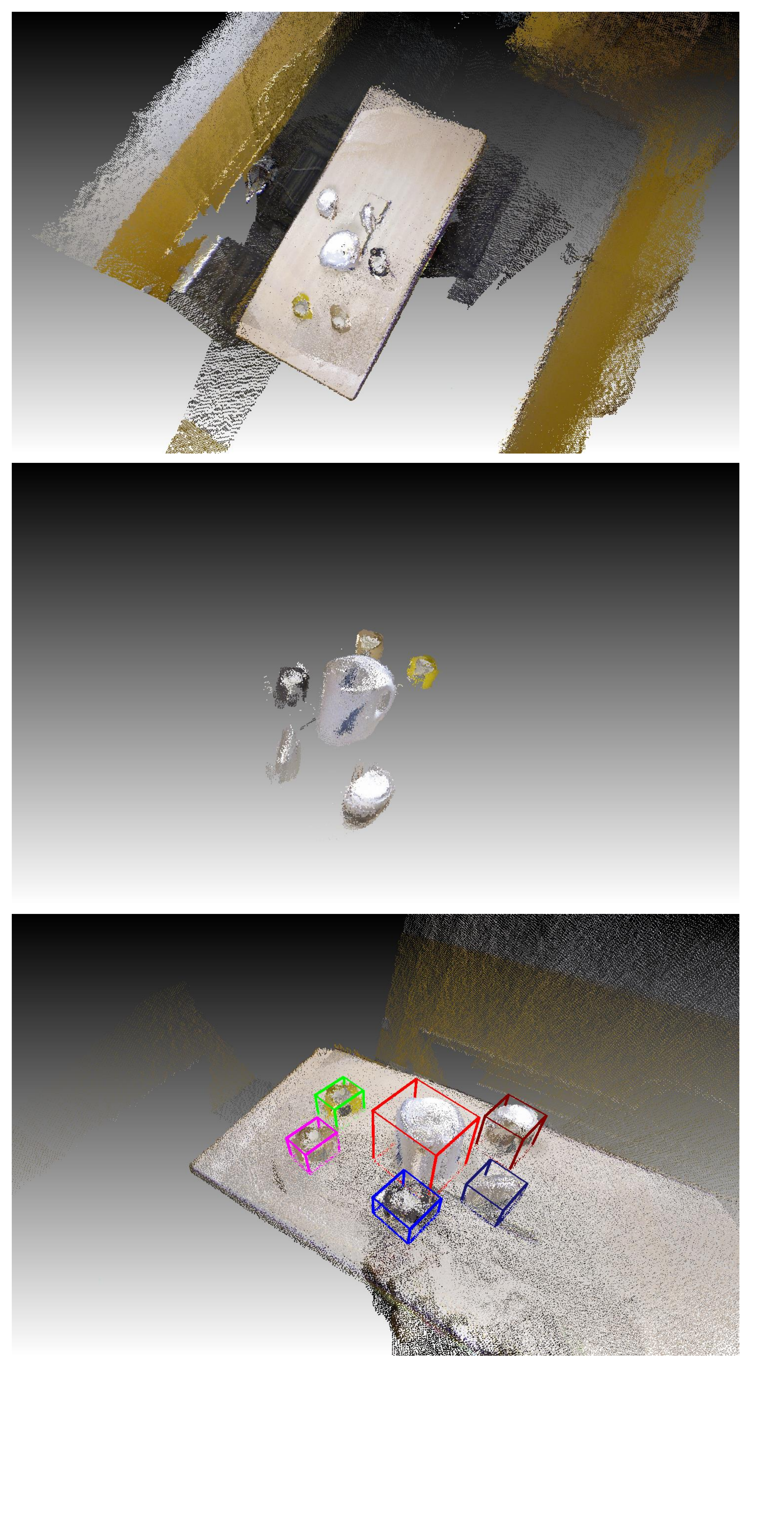} 
\caption{}
\end{subfigure} 
\caption{Our $3$D Object proposals for the UW-RGBD dataset (first three columns) and our dataset (last three columns). First row shows the point cloud of the different scenes. Second row shows our top ranked filtered points and finally third row highlights the resulting $3$D object proposals using density-based clustering. Our technique also picks up objects like desktop PC and chairs that are considered as background for most object segmentation algorithms.}
 \label{fig:downsampld_res}
\end{figure*}

\textbf{RGBD scene dataset 2011.} We use RGBD scenes as well to show our results in a more cluttered environment. RGBD scenes contain seven different scenes ranging from a lab to a kitchen. We ignore one scene - $desk_2$  as the camera pose obtained gave unsatisfactory results. We compare our results in $2$D with \cite{lai_icra2011} in Fig.~\ref{fig::rgbd_scenes}. We make an underlying assumption that our precise $3$D object proposals can be correctly identified as the relevant objects inside those bounding boxes. As the authors perform object detection, we manually measure the precision and recall for each object individually by using the ground-truth labeling per point as discussed previously. Our performance is significantly better than \cite{lai_icra2011}. We also observed that our technique picks up other objects such as laptop, computer mouse, and kitchen devices as objects of interest, which are treated as background in their analysis.

\begin{figure*}[t!]
	\begin{subfigure}[b]{0.245\textwidth}
	\includegraphics[width=\textwidth]{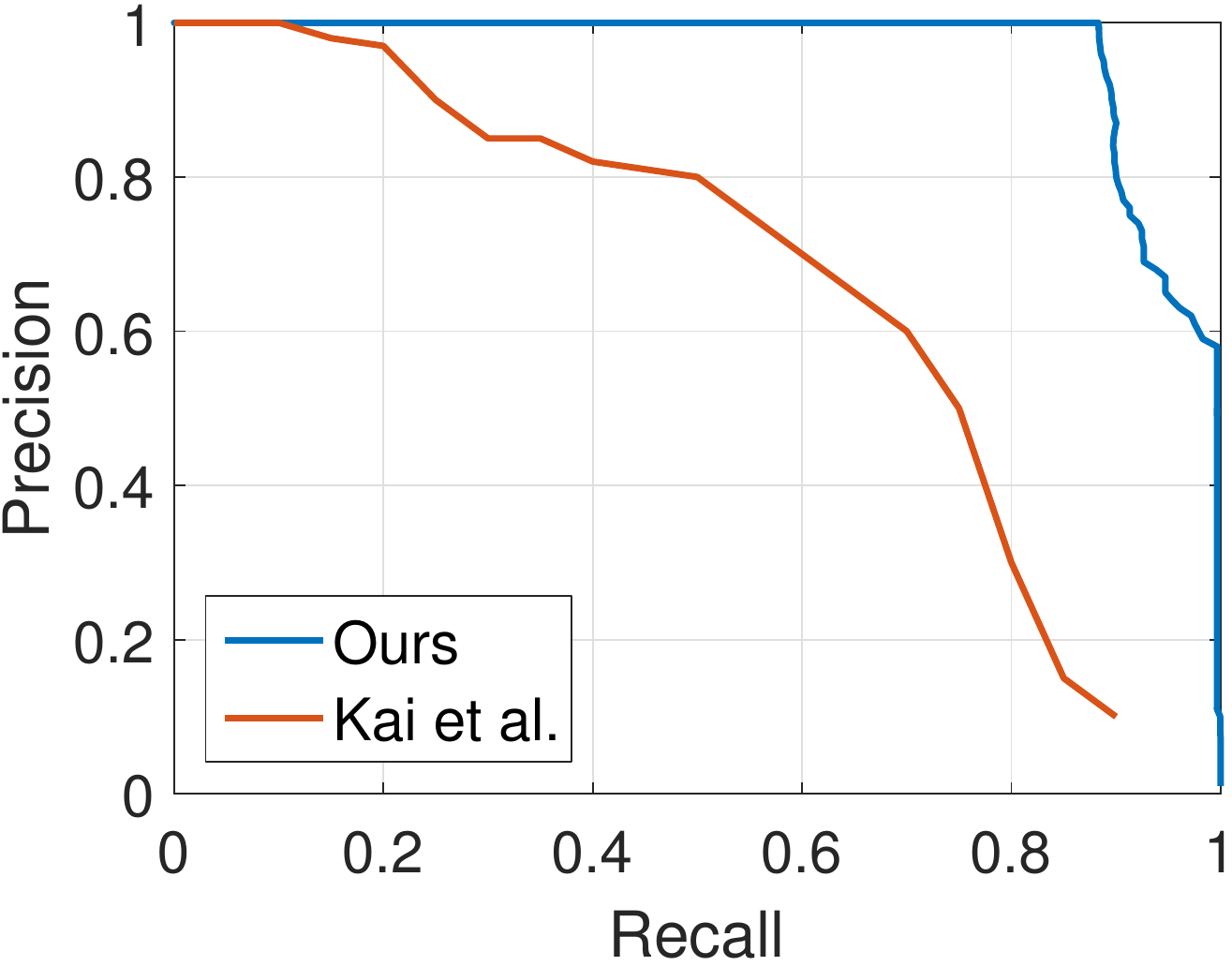}
	\caption{Bowl}
	\end{subfigure}
  \begin{subfigure}[b]{0.245\textwidth}
	\includegraphics[width=\textwidth]{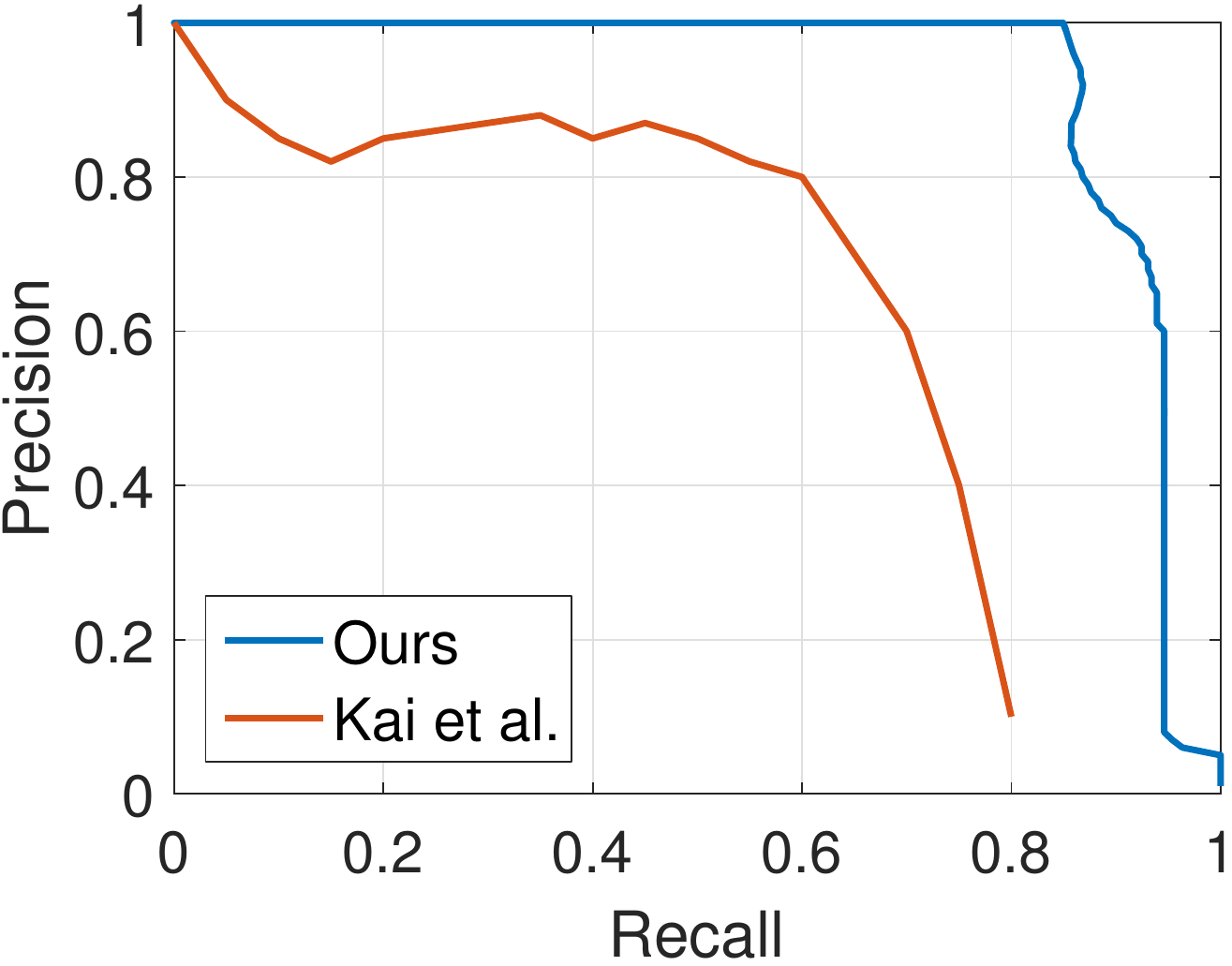}
	\caption{Cap}
	\end{subfigure}
  \begin{subfigure}[b]{0.245\textwidth}
	\includegraphics[width=\textwidth]{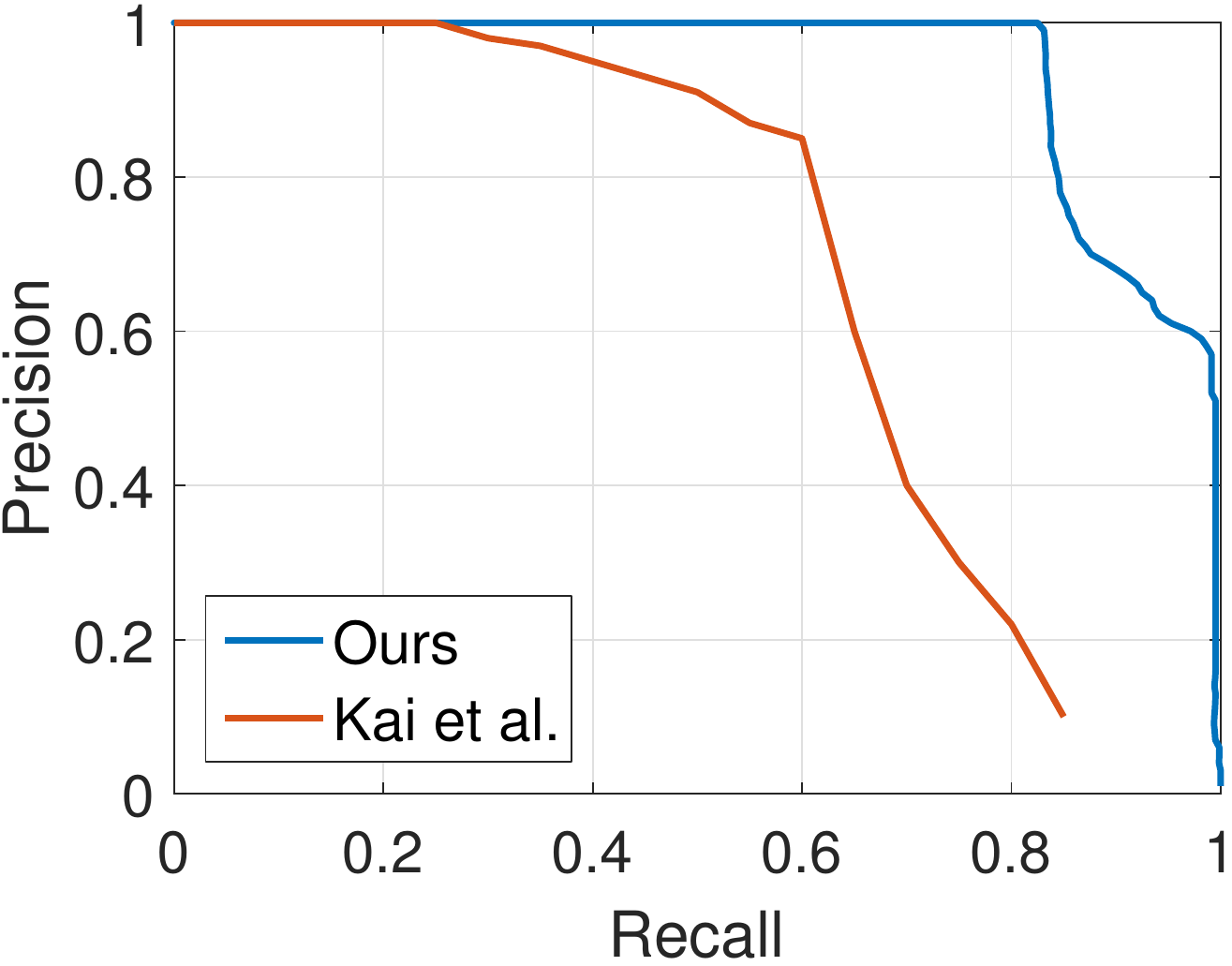}
	\caption{Coffee mug}
	\end{subfigure}
  \begin{subfigure}[b]{0.245\textwidth}
	\includegraphics[width=\textwidth]{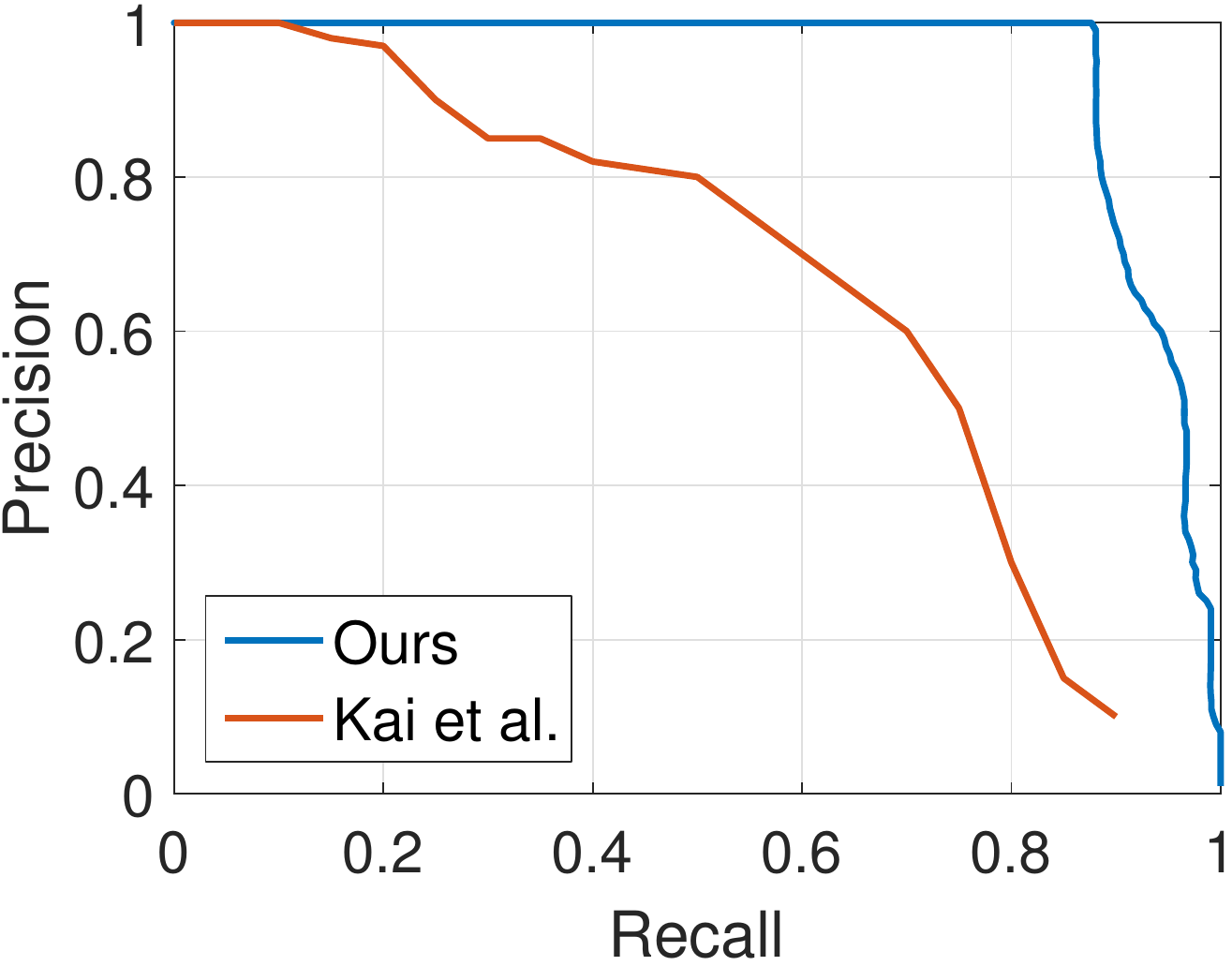}
	\caption{Soda can}
	\end{subfigure}
  \caption{We compare P-R measurements for the four objects in the RGBD scenes dataset 2011. We projected our $3$D bounding boxes onto the image plane to obtain our $2$D bounding boxes and used the ground-truth bounding boxes to obtain these results.}
 \label{fig::rgbd_scenes}
\end{figure*}

\textbf{Our dataset.} We collected our own dataset using Microsoft Kinect \textit{v}1. We captured six video sequences varying from 400 frames to 800 frames. Some of the scenes are shown in Fig.~\ref{fig:downsampld_res}. We placed multiple objects such as a kettle, books, coffee mugs and a messenger bag in various locations in an indoor room. We purposely placed these objects close to each other to showcase that our approach can cluster these nearby objects separately. We also placed coffee mugs on a whiteboard attached to the wall (second scene from right) to highlight that our approach is also able to recognize them as objects of interests despite not being placed on the table. 

Our approach is able to identify small objects such as power bank lying on the table consistently. The plane removal step assists in segmenting out the books placed flat on the table. This would be extremely difficult to do in the global point cloud where the table $3$D points may not lie on a unit plane due to depth noise and SLAM errors. However, due to the presence of depth noise, some table pixels are not filtered out, and become a part of the nearby objects. This leads to clustering two neighboring objects together as seen in Fig.~\ref{fig:downsampld_res}(e), where the power bank and books are clustered together as one object. 

\begin{table}
\resizebox{1.0\columnwidth}{!}{ \begin{tabular}{ |c|c|c|c|c| }
\hline
Method & Run-time & AP & AR & F-measure \\ \hline
DetOnly \cite{lai2012detection} & $1.8$s & $61.7$ & $81.9$ & $70.38$ \\ \hline
HMP2D+3D \cite{lai_icra14} & $4.0$s & $92.8$ & $\textbf{95.3}$ & $\textbf{94.03}$ \\ \hline
Pillai \& Leonard \cite{pillai2015monocular} & $1.6$s & $81.5$ & $59.4$ & $68.72 $ \\\hline
Ours    & $3.03$s & ${93.46}$ & {$76.19$} & {$83.95$}\\ \hline
Ours w$/$o chairs,desktop   & $3.03$s & $\textbf{98.76}$ & $81.66$ & $89.40$ \\ \hline
Ours($\downarrow 2$) & $\textbf{0.97}$s & $91.91$ & $78.49$& $84.67$\\ \hline
Ours($\downarrow 2$) w$/$o chairs,desktop & ${0.97}$s  & $96.95$ & $82.86$ & $89.35$ \\ \hline
\end{tabular}
 }
 \caption{Analysis of our $3$D object proposals on the UW-RGBD scene dataset in comparison to \cite{lai2012detection},\cite{lai_icra14}, and \cite{pillai2015monocular}. We achieve efficient run-time performance if we downsample the data for our analysis. Our experiments were conducted on a single core Intel Xeon E5-1620 CPU.} 
\label{table::time_comp}
\end{table}

\end{subsection}

\end{section}

\begin{section}{Conclusion}\label{sec:conclusion}

In this paper, we have developed a novel multi-view based $3$D object proposal technique using depth information along with initial $2$D proposals. To our knowledge, this is the first technique that truly produces $3$D object proposals without using trained object segmentation and recognition classifiers. In future work, we aim to optimize our system towards real-time $3$D object proposals over even larger environments by exploring multi-scale representations for memory and computational efficiency. Ultimately, we intend to integrate our system with SLAM to improve its accuracy by treating the object proposals as fixed landmarks in the scene.
\end{section}

\ifCLASSOPTIONcaptionsoff
  \newpage
\fi

{
\bibliographystyle{IEEEtran}
\bibliography{SLAM-O1}
}

%

\begin{IEEEbiography}[{\includegraphics[width=1in,height=1.25in,clip,keepaspectratio]{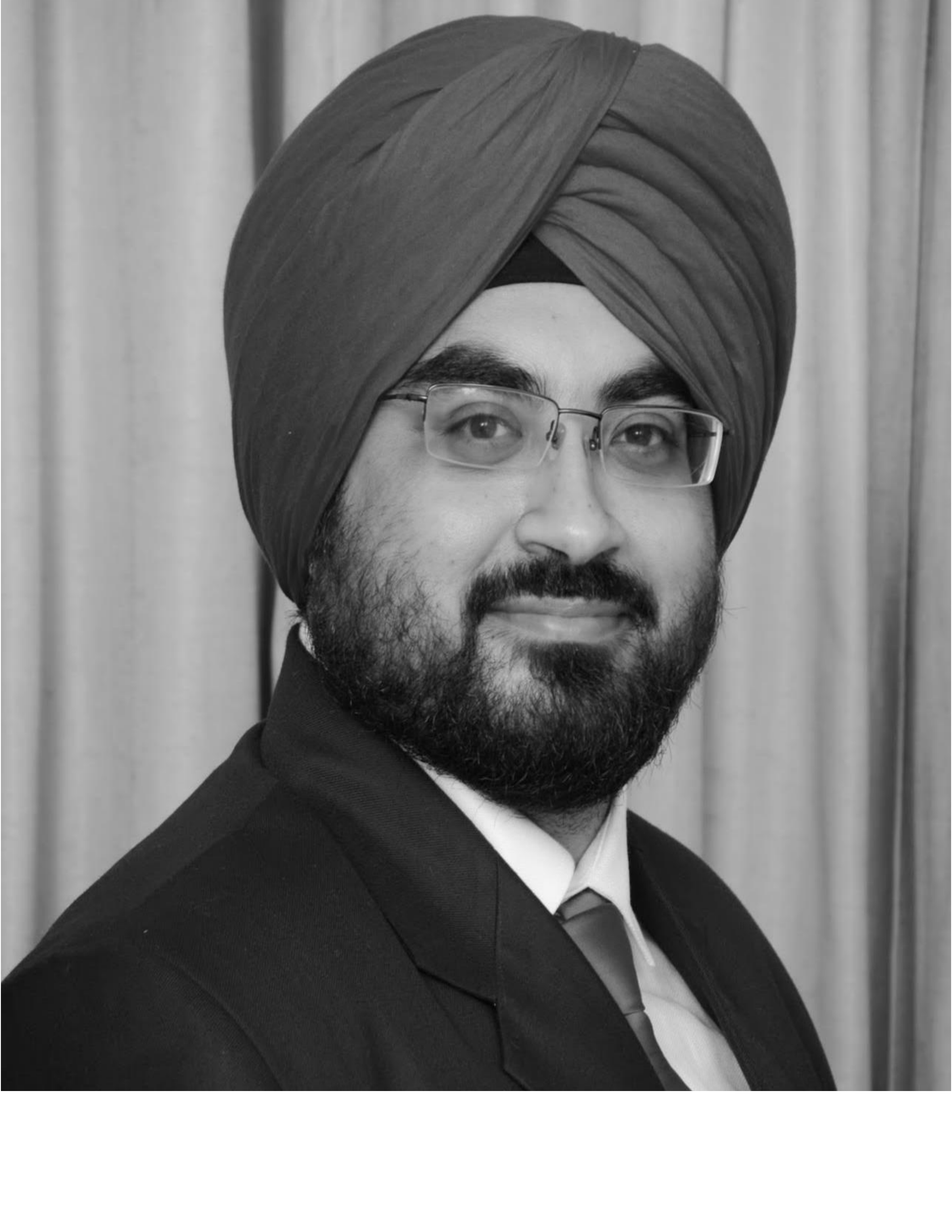}}]
{Ramanpreet Singh Pahwa} received his B.Sc. from Carnegie Mellon University, USA in 2007. He worked in Data Storage Institute and Institute for Infocomm Research, Singapore from 2007 to 2009. He is currently pursuing his Ph.D. in University of Illinois at Urbana-Champaign (UIUC), USA and working as a summer intern in  Advanced Digital Sciences Center (ADSC), Singapore. 

His research interests include computer and robot vision, 3D reconstruction, and depth cameras. 
\end{IEEEbiography}

\begin{IEEEbiography}[{\includegraphics[width=1in,height=1.25in,clip,keepaspectratio]{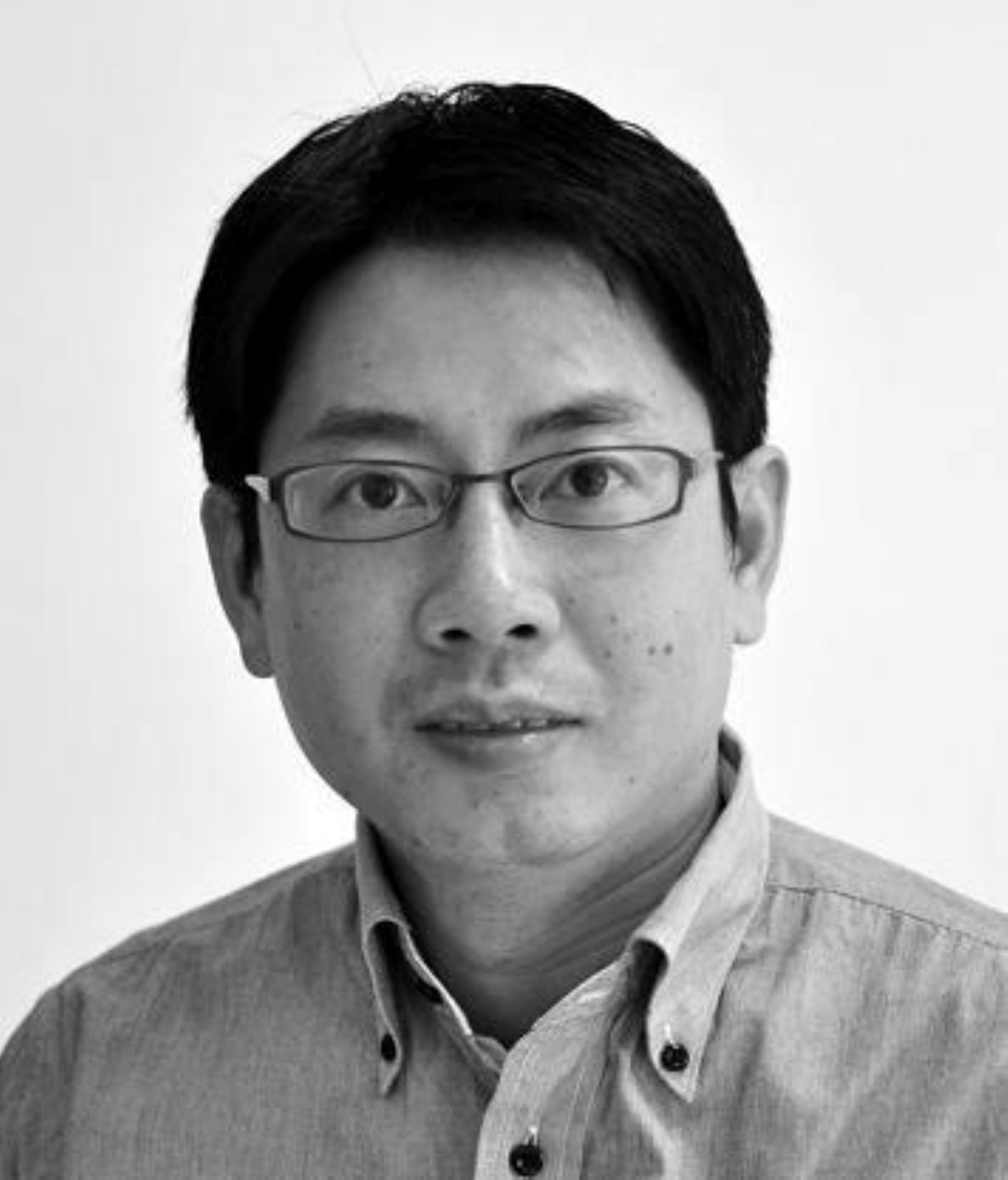}}]{Jiangbo Lu} (M'09-SM'15) received his B.S. and M.S. degrees in electrical engineering from Zhejiang University, Hangzhou, China, in 2000 and 2003, respectively, and the Ph.D. degree in electrical engineering, Katholieke Universiteit Leuven, Leuven, Belgium, in 2009.

Since September 2009, he has been working with the Advanced Digital Sciences Center, Singapore, which is a joint research center between the University of Illinois at Urbana-Champaign, USA, and the Agency for Science, Technology and Research (A*STAR), Singapore, where he is leading a few research projects as a Senior Research Scientist. His research interests include computer vision, visual computing, image processing, video communication, interactive multimedia applications and systems, and efficient algorithms for various architectures.

Dr. Lu served as an Associate Editor for IEEE Transactions on Circuits and Systems for Video Technology (TCSVT) in 2012-2016. He received the 2012 TCSVT Best Associate Editor Award.
\end{IEEEbiography}
\vspace{-1.1cm}

\begin{IEEEbiography}[{\includegraphics[width=1in,height=1.25in,clip,keepaspectratio]{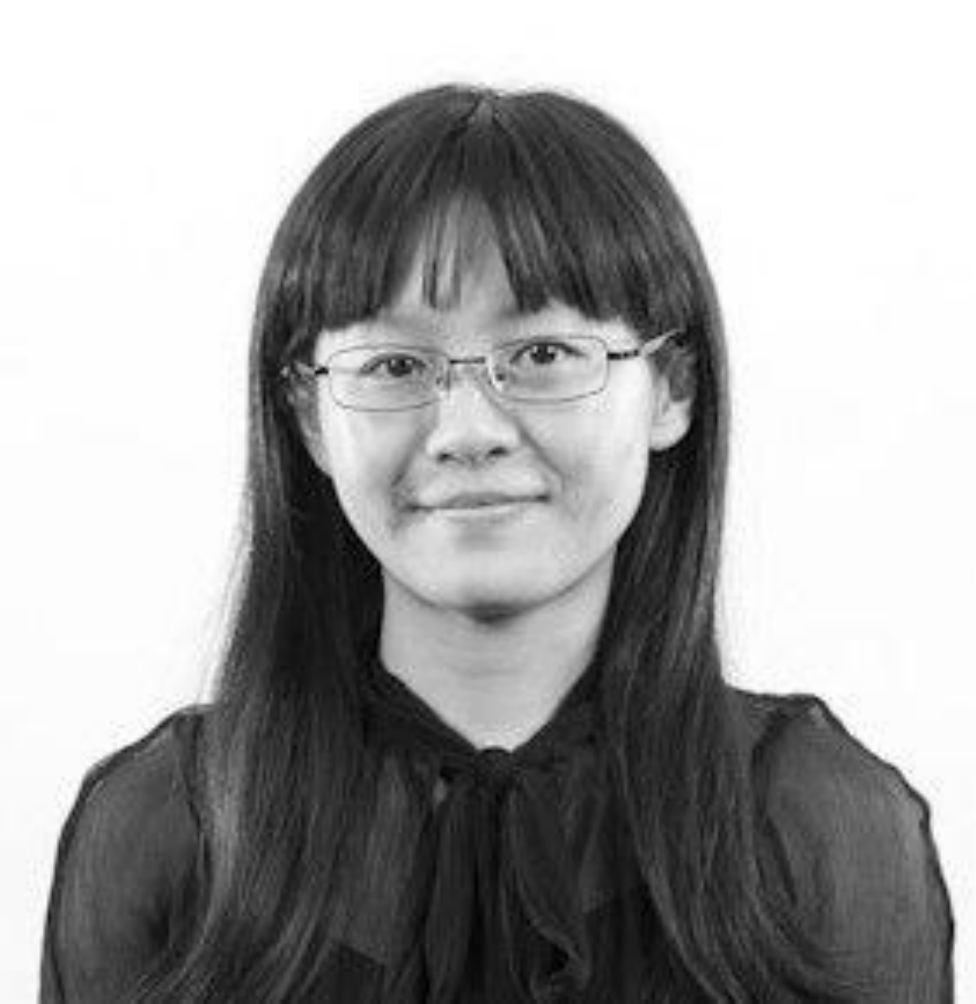}}]
{Nianjuan Jiang} received her B.E. and Ph.D. degree from Department of Electrical $\&$ Computer Engineering, National University of Singapore in 2007 and 2013, respectively. She is currently a post-doc researcher with ADSC. 

Her research interest includes computer vision and computer graphics, and especially efficient and robust 3D reconstruction systems.
\end{IEEEbiography}

\vspace{-1.1cm}
\begin{IEEEbiography}[{\includegraphics[width=1in,height=1.25in,clip,keepaspectratio]{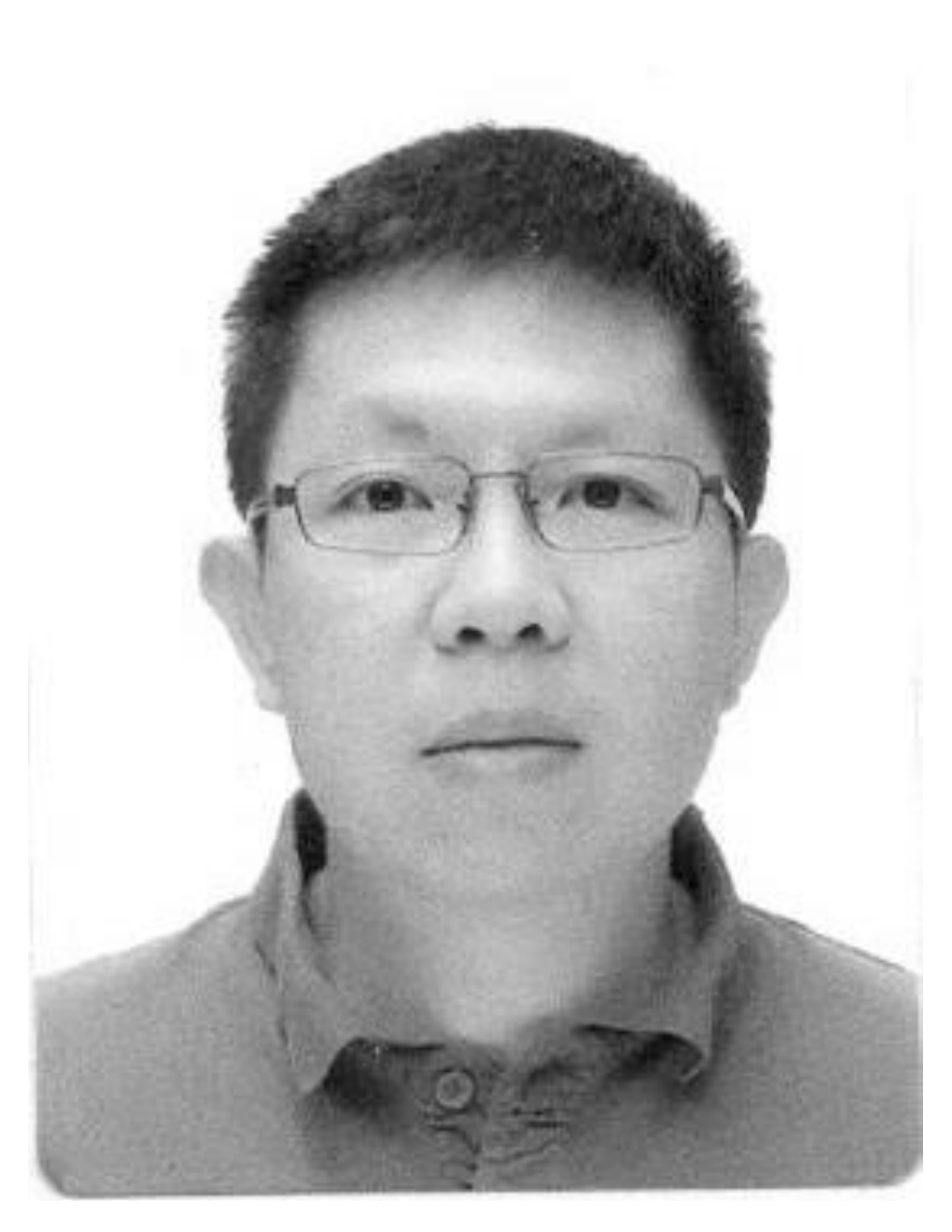}}]
{Tian Tsong Ng} is a research staff in Institute for Infocomm Research, Singapore, since 2007 and currently a deputy department head for Situational Awareness Analytics Department. He received his B.Eng from Malaya University in 1998, M.Phil. from Cambridge University in 2001, and Ph.D. from Columbia University in 2007. 

He won the Microsoft Best Student Paper Award at ACM Multimedia Conference in 2005, the John Wiley $\&$ Sons Best Paper Award at the IEEE Workshop in Information Security and Forensics in 2009.
\end{IEEEbiography}

\vspace{-1.1cm}
\begin{IEEEbiography}[{\includegraphics[width=1in,height=1.25in,clip,keepaspectratio]{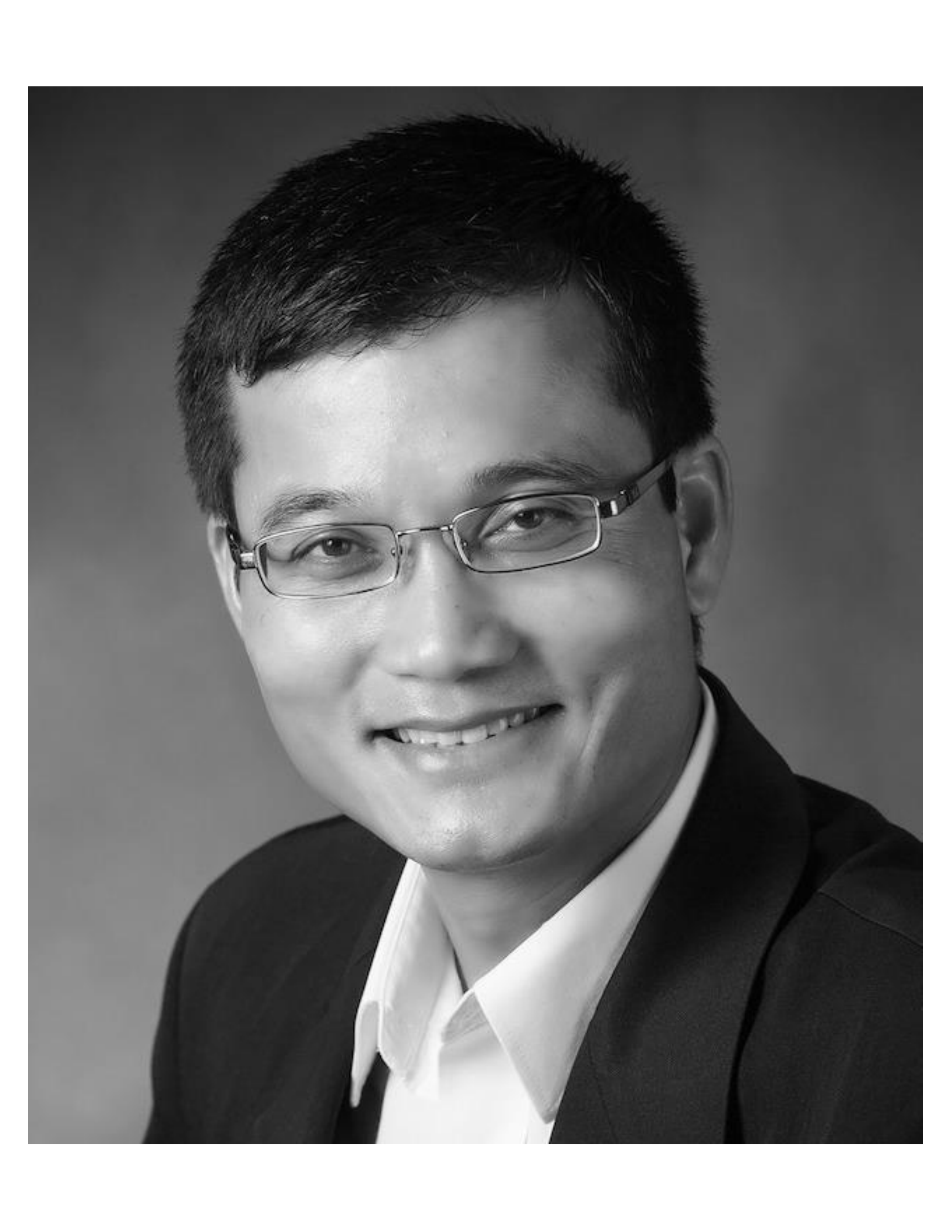}}]
{Minh N. Do} (M'01, SM'07, F'14) received the B.Eng. degree in computer engineering from the University of Canberra, Australia, in 1997, and the Dr.Sci. degree in communication systems from the Swiss Federal Institute of Technology Lausanne (EPFL), Switzerland, in 2001.

Since 2002, he has been on the faculty at the University of Illinois at Urbana-Champaign (UIUC), where he is currently a Professor in the Department of ECE, and holds joint appointments with the Coordinated Science Laboratory, the Beckman Institute for Advanced Science and Technology, and the Department of Bioengineering.  His research interests include signal processing, computational imaging, geometric vision, and data analytics.

He received a CAREER Award from the National Science Foundation in 2003, and a Young Author Best Paper Award from IEEE in 2008.  He was named a Beckman Fellow at the Center for Advanced Study, UIUC, in 2006, and received of a Xerox Award for Faculty Research from the College of Engineering, UIUC, in 2007.  He was a member of the IEEE Signal Processing Theory and Methods Technical Committee, Image, Video, and Multidimensional Signal Processing Technical Committee, and an Associate Editor of the IEEE Transactions on Image Processing.  
\end{IEEEbiography}
\end{document}